\documentclass[a4paper,fleqn]{cas-dc}

\usepackage[authoryear]{natbib}
\usepackage[T1]{fontenc}
\usepackage{adjustbox}   

\usepackage{subcaption}
\usepackage{tikz}
\usetikzlibrary{decorations.pathreplacing, arrows.meta}
\usetikzlibrary{positioning, fit, backgrounds, calc, shapes.geometric}
\usepackage{soul}
\usepackage{moresize}
\usepackage{mathtools}
\usepackage{fixmath}
\usepackage{enumitem}

\usepackage{CJKutf8}

\definecolor{figJoint}{RGB}{209,231,205}     
\definecolor{figIntent}{RGB}{252,224,196}    
\definecolor{figSlot}{RGB}{219,209,236}      
\definecolor{figSys}{RGB}{237,237,237}       
\definecolor{figUsr}{RGB}{206,226,242}       

\definecolor{darkblue}{rgb}{0, 0, 0.5}
\hypersetup{colorlinks=true,citecolor=darkblue, linkcolor=darkblue, urlcolor=darkblue}

\definecolor{aquamarine}{rgb}{0.5, 1.0, 0.83}
\definecolor{babypink}{rgb}{0.96, 0.76, 0.76}
\definecolor{flavescent}{rgb}{0.97, 0.91, 0.56}
\definecolor{pastelorange}{rgb}{1.0, 0.7, 0.28}
\definecolor{turquoiseblue}{rgb}{0.0, 1.0, 0.94}
\definecolor{junebud}{rgb}{0.74, 0.85, 0.34}
\definecolor{mauve}{rgb}{0.88, 0.69, 1.0}
\definecolor{olivine}{rgb}{0.6, 0.73, 0.45}
\definecolor{paleaqua}{rgb}{0.74, 0.83, 0.9}
\definecolor{palegreen}{rgb}{0.6, 0.98, 0.6}
\definecolor{palespringbud}{rgb}{0.93, 0.92, 0.74}

\DeclareRobustCommand{\hlaquamarine}[1]{{\sethlcolor{aquamarine}\hl{#1}}}

\DeclareRobustCommand{\hlflavescent}[1]{{\sethlcolor{flavescent}\hl{#1}}}
\DeclareRobustCommand{\hlpastelorange}[1]{{\sethlcolor{pastelorange}\hl{#1}}}
\DeclareRobustCommand{\hlturquoiseblue}[1]{{\sethlcolor{turquoiseblue}\hl{#1}}}
\DeclareRobustCommand{\hljunebud}[1]{{\sethlcolor{junebud}\hl{#1}}}
\DeclareRobustCommand{\hlmauve}[1]{{\sethlcolor{mauve}\hl{#1}}}
\DeclareRobustCommand{\hlolivine}[1]{{\sethlcolor{olivine}\hl{#1}}}
\DeclareRobustCommand{\hlpaleaqua}[1]{{\sethlcolor{paleaqua}\hl{#1}}}
\DeclareRobustCommand{\hlpalegreen}[1]{{\sethlcolor{palegreen}\hl{#1}}}

\DeclareRobustCommand{\atisFromLoc}[1]{\underbracket{\text{\hlaquamarine{#1}}}_{\texttt{from\_loc.city\_name}}}
\DeclareRobustCommand{\atisToLoc}[1]{\underbracket{\text{\hlflavescent{#1}}}_{\texttt{to\_loc.city\_name}}}
\DeclareRobustCommand{\atisAirport}[1]{\underbracket{\text{\hlpastelorange{#1}}}_{\texttt{airport\_name}}}
\DeclareRobustCommand{\atisCostRelative}[1]{\underbracket{\text{\hlturquoiseblue{#1}}}_{\texttt{cost\_relative}}}
\DeclareRobustCommand{\atisCityName}[1]{\underbracket{\text{\hljunebud{#1}}}_{\texttt{city\_name}}}
\DeclareRobustCommand{\atisDepartTimePeriodOfDay}[1]{\underbracket{\text{\hlmauve{#1}}}_{\texttt{depart\_time.period\_of\_day}}}
\DeclareRobustCommand{\atisAirlineCode}[1]{\underbracket{\text{\hlolivine{#1}}}_{\texttt{airline\_code}}}
\DeclareRobustCommand{\atisDepartTime}[1]{\underbracket{\text{\hlpaleaqua{#1}}}_{\texttt{depart\_time.time}}}
\DeclareRobustCommand{\atisAirportCode}[1]{\underbracket{\text{\hlpalegreen{#1}}}_{\texttt{airport\_code}}}

\providecommand{\impact}[1]{\section*{Broader Impact Statement}#1}
\providecommand{\acks}[1]{\section*{Acknowledgements}#1}

\begin{document}

\let\WriteBookmarks\relax
\let\printorcid\relax   

\shorttitle{A Survey of Intent Classification and Slot-Filling Datasets}
\shortauthors{S. Larson and K. Leach}

\title[mode=title]{A Survey of Intent Classification and Slot-Filling Datasets for Task-Oriented Dialog}

\author[1]{Stefan Larson}
\cormark[1]
\ead{stefan.larson@vanderbilt.edu}

\author[1]{Kevin Leach}
\ead{kevin.leach@vanderbilt.edu}

\cortext[1]{Corresponding author}

\affiliation[1]{organization={Department of Computer Science, Vanderbilt University},
            addressline={1400 18th St. S},
            city={Nashville},
            postcode={37212},
            state={TN},
            country={USA}}

\begin{abstract}
Interest in dialog systems has grown substantially in the past decade. By extension, so too has interest in developing and improving intent classification and slot-filling models, which are two components that are commonly used in task-oriented dialog systems. Moreover, good evaluation benchmarks are important in helping to compare and analyze systems that incorporate such models. Unfortunately, much of the literature in the field is limited to analysis of relatively few benchmark datasets. In an effort to promote more robust analyses of task-oriented dialog systems, we have conducted a survey of publicly available datasets for the tasks of intent classification and slot-filling. We catalog the important characteristics of each dataset, and offer discussion on the applicability, strengths, and weaknesses of each. The emergence of large language models (LLMs) as capable zero-shot NLU systems gives such benchmarks a new role: the corpora cataloged here provide the principled evaluation infrastructure needed to measure LLM capabilities in structured NLU tasks, compare them against specialized models, and identify the settings---multilingual, multi-intent, low-resource---where significant gaps remain. Our goal is that this survey aids in increasing the accessibility of these datasets, which we hope will enable their use in future evaluations of intent classification and slot-filling models, whether those models are task-specific classifiers, fine-tuned language models, or zero-shot LLMs.
\end{abstract}

\begin{keywords}
intent classification \sep slot-filling \sep task-oriented dialog \sep datasets \sep natural language understanding
\end{keywords}

\maketitle

\section{Introduction}
Task-oriented dialog systems are among the most widely deployed applications of Natural Language Processing (NLP), powering voice assistants, in-app support agents, and customer service chatbots.
Within the academic research community, however, task-oriented dialog system models are often benchmarked on relatively few evaluation datasets.
This is in spite of the fact that the past few years have seen a substantial growth in the number of available datasets for building and evaluating intent classification and slot-filling models for task-oriented dialog systems.
Thus, the goal of this survey is to catalog these intent classification and slot-filling datasets to help facilitate their use in building and evaluating dialog systems and beyond.

Other surveys have discussed dialog datasets in depth \citep{serban2018survey}, but exclude almost all intent classification and slot-filling datasets,
and model-focused surveys on dialog systems focus mainly on models and pay much less attention to datasets.
In this paper, we instead emphasize datasets themselves and present a survey of 50 corpora for developing and evaluating intent classification and slot-filling models.\footnote{We count the corpora cataloged in Tables~\ref{tab:joint-datasets}, \ref{tab:other-datasets}, \ref{tab:intent-classification-datasets}, and \ref{tab:slot_filling_datasets}, counting the constituent datasets of collections such as the \emph{Braun Collection}, the \emph{MIT Collection}, the \emph{Jaech Collection}, and \emph{HINT3} individually.}
Where appropriate, we discuss the strengths and weaknesses of each dataset, and highlight points of uniqueness for each. 

The arrival of large language models (LLMs) as capable zero-shot natural language understanding systems has raised a fair question: are task-specific intent classification and slot-filling benchmarks still relevant?
We argue that they are.
Precisely because LLMs can be prompted to perform these tasks without any task-specific training, the community now has both the opportunity and the responsibility to evaluate those capabilities rigorously against principled, human-labeled benchmarks.
Without such benchmarks, claims about LLM performance in structured NLU tasks rest on informal demonstrations rather than reproducible evidence.
The datasets cataloged in this survey provide exactly that infrastructure: held-out test sets with gold-standard labels, covering diverse domains, languages, and annotation schemes, against which any model---task-specific or general-purpose---can be fairly compared.

This survey paper consists of the following:
In Section~\ref{sec:related-work}, we briefly discuss prior surveys related to task-oriented dialog and datasets.
Section~\ref{sec:dialog-systems} provides an overview of task-oriented dialog, intent classification, and slot-filling, as well as a discussion of various utterance types, data sources for constructing datasets, and the metrics used to evaluate models on them (Section~\ref{sec:evaluation-metrics}).
Sections~\ref{sec:joint-datasets}, \ref{sec:intent-classification-datasets}, \ref{sec:slot-filling-datasets}, and \ref{sec:other-datasets} are dedicated to introducing and discussing corpora:
specifically, joint intent classification and slot-filling\footnote{Also referred to as ``joint modeling''.} (Section~\ref{sec:joint-datasets}), intent classification (Section~\ref{sec:intent-classification-datasets}), slot-filling (Section~\ref{sec:slot-filling-datasets}), and other related corpora (Section~\ref{sec:other-datasets}).
Figure~\ref{fig:datasets-surveyed} presents an overview of the datasets surveyed in Sections~\ref{sec:joint-datasets}--\ref{sec:slot-filling-datasets}.
We distinguish corpora according to several factors: size of the data, problem solved by the dataset, and provenance of the dataset.
We organize our survey of datasets by dataset task (i.e., joint modeling, intent classification, and slot-filling), as opposed to---for instance---application domain (banking, travel, etc.), as this task typology closely matches how these datasets are consumed by researchers.
However, we also mention dataset domains where appropriate; a visual guide of datasets surveyed in this paper broken down by model task and domain is displayed in Figure~\ref{fig:datasets-surveyed}.
Finally, Section~\ref{sec:discussion-future-work} provides a discussion of the landscape of the surveyed corpora, with commentary on successes, pitfalls, and lessons learned from the datasets.  
We also discuss remaining challenge areas that are not currently covered by existing corpora to encourage the development of future research and evaluation of new datasets in this domain.

\begin{figure*}[p]
\centering
\adjustbox{max totalsize={\textwidth}{0.8\textheight}}{%
\begin{tikzpicture}[
  font=\footnotesize,
  cell/.style={align=center, text width=4.0cm, font=\footnotesize\itshape},
  rowlbl/.style={align=center, text width=2.0cm, font=\footnotesize\bfseries},
  hdr/.style={rounded corners=4pt, draw=black!55, thick,
              inner xsep=3mm, inner ysep=0pt},
  panel/.style={rounded corners=8pt, draw=black!55, thick},
]
\def\colA{4.75} \def\colB{9.25} \def\colC{13.75}
\def\panw{2.15}
\def\panTop{-0.9} \def\panBot{-20.4}

\newcommand{\hdrtext}[1]{\begin{minipage}[c][1.75cm][c]{3.8cm}%
  \centering\small\bfseries #1\end{minipage}}

\fill[figJoint,  panel] (\colA-\panw, \panTop) rectangle (\colA+\panw, \panBot);
\fill[figIntent, panel] (\colB-\panw, \panTop) rectangle (\colB+\panw, \panBot);
\fill[figSlot,   panel] (\colC-\panw, \panTop) rectangle (\colC+\panw, \panBot);
\draw[panel] (\colA-\panw, \panTop) rectangle (\colA+\panw, \panBot);
\draw[panel] (\colB-\panw, \panTop) rectangle (\colB+\panw, \panBot);
\draw[panel] (\colC-\panw, \panTop) rectangle (\colC+\panw, \panBot);

\node[hdr, fill=figJoint]  at (\colA, 0.1) {\hdrtext{Section 5:\\Joint Datasets}};
\node[hdr, fill=figIntent] at (\colB, 0.1) {\hdrtext{Section 6:\\Intent Classification Datasets}};
\node[hdr, fill=figSlot]   at (\colC, 0.1) {\hdrtext{Section 7:\\Slot-Filling Datasets}};

\foreach \y in {-1.0, -3.9, -6.0, -7.6, -9.4, -11.8, -14.2, -18.6, -20.3}
  \draw[densely dotted, black!60] (-0.2, \y) -- (16.1, \y);

\node[rowlbl] at (1.1, -2.45) {travel \& hotels};
\node[cell] at (\colA, -2.45) {ATIS, \underline{Multilingual ATIS},\\
  \underline{MultiATIS++}, MixATIS\\ \underline{PhoATIS}, \underline{Persian ATIS},\\ NLU++, HWU-64};
\node[cell] at (\colB, -2.45) {Clinc-150, Redwood};
\node[cell] at (\colC, -2.45) {Jaech Airbnb, Jaech Greyhound,\\ Jaech United};

\node[rowlbl] at (1.1, -4.95) {banking};
\node[cell] at (\colA, -4.95) {NLU++};
\node[cell] at (\colB, -4.95) {Clinc-150, Outlier,\\ Banking-77, Redwood,\\ \underline{MInDS-14}};

\node[rowlbl] at (1.1, -6.8) {insurance};
\node[cell] at (\colB, -6.8) {ACID, Clinc-150,\\ Redwood};

\node[rowlbl] at (1.1, -8.5) {restaurants \&\\ movies};
\node[cell] at (\colA, -8.5) {Snips, MixSNIPS,\\ \underline{Almawave-SLU}, HWU-64};
\node[cell] at (\colB, -8.5) {Clinc-150, Redwood};
\node[cell] at (\colC, -8.5) {MIT Movie, MIT Restaurant, Jaech\\ OpenTable, Restaurants-8k};

\node[rowlbl] at (1.1, -10.6) {home automation};
\node[cell] at (\colA, -10.6) {Snips, MixSNIPS, \underline{Almawave-SLU},\\
  Fluent Speech Commands,\\ \underline{Leyzer}, HWU-64};
\node[cell] at (\colB, -10.6) {Clinc-150, Redwood};

\node[rowlbl] at (1.1, -13.0) {e-commerce, sales\\ \& support};
\node[cell] at (\colA, -13.0) {Braun askUbuntu,\\ Braun webApplications};
\node[cell] at (\colB, -13.0) {HINT3 SOFMattress,\\ HINT3 Curekart,\\ HINT3 Powerplay11,\\ CREDIT16};

\node[rowlbl] at (1.1, -16.4) {general};
\node[cell] at (\colA, -16.4) {Snips, MixSNIPS, \underline{Almawave-SLU},\\
  TOP, \underline{Leyzer}, \underline{CSTOP}, \underline{MTOP}\\
  Braun chatbot, HWU-64, \underline{CSTOP},\\
  TOPv2, \underline{Facebook}, \underline{FewJoint},\\ \underline{CAIS}, \underline{xSID},
  \underline{MASSIVE}, BlendX};
\node[cell] at (\colB, -16.4) {Clinc-150, Redwood,\\ \underline{SMP-ECDT Task 1},\\ ROSTD, StanfordLU};

\node[rowlbl] at (1.1, -19.45) {food ordering};
\node[cell] at (\colC, -19.45) {Pizza, FoodOrdering (Coffee,\\ Burrito, Burger, Sub)};
\end{tikzpicture}}
    \caption{Outline of datasets surveyed, grouped by model task type (vertical bars) and by topic domain (horizontal bars). Datasets underlined contain non-English utterances.}
    \label{fig:datasets-surveyed}
\end{figure*}


\section{Related Surveys}\label{sec:related-work}

In this section, we discuss areas of research already addressed by previous surveys to place our own work in context.  
Existing surveys have addressed (1) dialog systems \citep{chen-2017-survey, gao2019neural, mctear-book, Ni2021RecentAI, yi-2026-llm-multiturn-survey}, 
(2) task-oriented dialog management \citep{dialog-management-survey},
(3) language understanding such as intent classification and slot-filling models \citep{Korpusik2019,Hou_2019, recent-advances-survey, Liu_2019, louvan-magnini-2020-recent, slu-survey, razumovskaia2021crossing, weld2021survey},
and (4) semantic parsing \citep{semantic-parsing-survey}.
However, these surveys do not thoroughly discuss evaluation corpora (e.g., discussion is often limited to \emph{ATIS} and \emph{Snips} in the case of \citet{weld2021survey}'s survey). 
While \citet{serban2018survey} is an exception with respect to datasets and benchmarks, their survey focuses mostly on end-to-end dialog corpora and does not discuss task-oriented dialog datasets aside from \emph{ATIS}.

Prior work has also compiled collections of evaluation datasets for task-oriented dialog. This work includes \citet{Henderson2019}, \citet{larson-etal-2020-data}, and \citet{MehriDialoGLUE2020}, but none of these are nearly as broad as our survey.
\citet{deriu:hal-03006231} surveys evaluation \emph{methods} for dialog systems, but focuses mostly on dialog paradigms outside of intent classification and slot-filling (e.g., open-ended dialog). 
Prior surveys on building or constructing datasets are limited, but include \citet{mohammad-survey}, which surveys data acquisition methods for building ``chatbot''-style task-oriented systems. 
Surveys on datasets outside of dialog systems include those for machine reading comprehension \citep{Dzendzik2021EnglishMR}, explainable NLP \citep{wiegreffe2021teach}, and question answering \citep{qa-survey-2022}. 
Finally, intent classification and slot-filling datasets are often used to evaluate few- and zero-shot learning algorithms, and surveys on this subject include \citet{few-shot-survey} and \citet{yin2020metalearning}.
This paper thus fills a gap in the survey literature on dialog systems by surveying and cataloging intent classification and slot-filling datasets.


\begin{figure*}[t]
\centering
\includegraphics[width=1\textwidth]{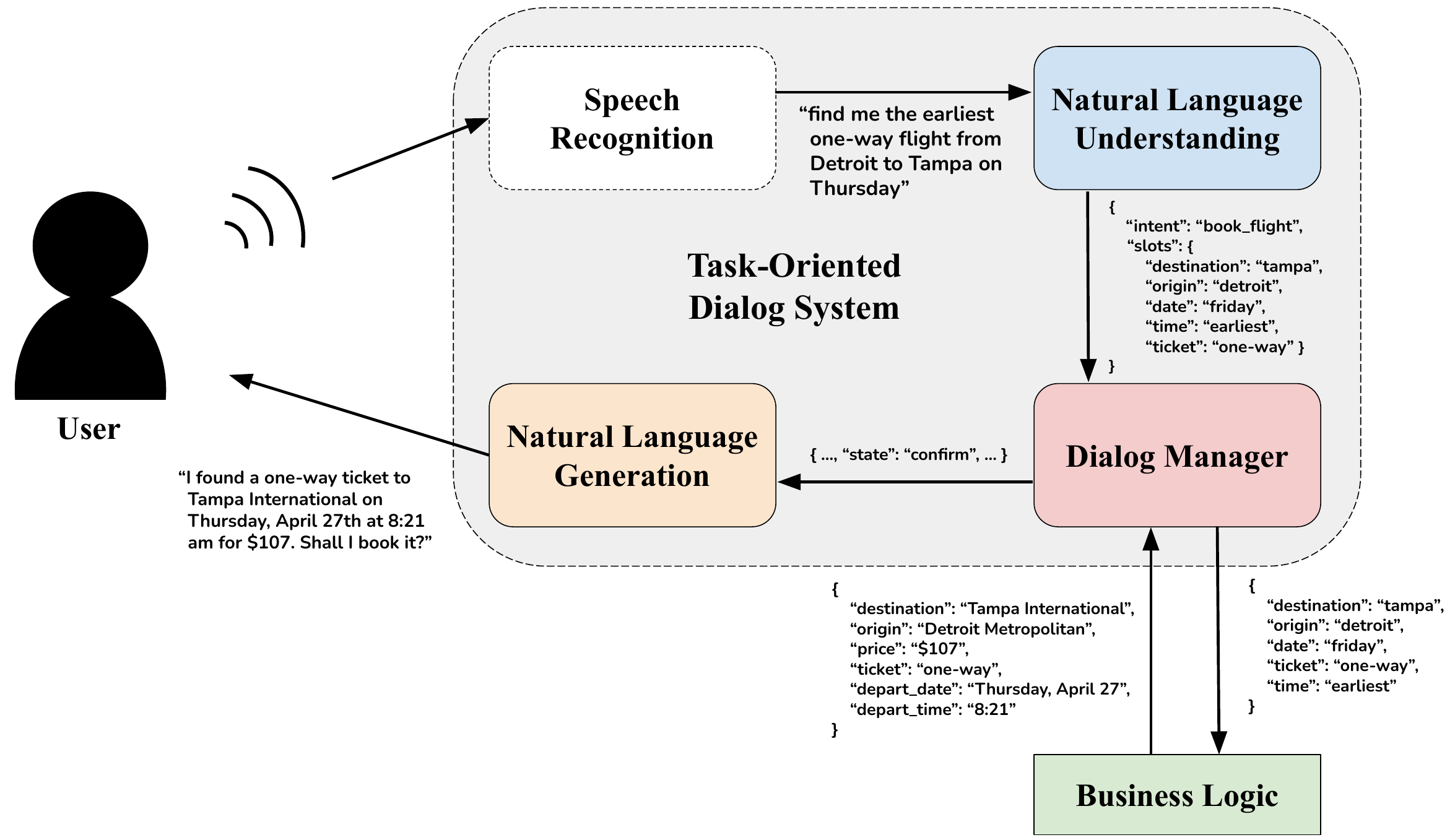}
\caption{Diagram of typical task-oriented dialog systems. A user's utterance is first transcribed using speech recognition or is provided directly as textual input from the user.
The Natural Language Understanding (NLU) module then extracts semantically meaningful information from the query. In this survey, our focus is on building and evaluating datasets that allow NLU modules to train intent classification and slot-filling models. The extracted information from the NLU module is passed to the Dialog Manager, which may interact with ``backend'' Business Logic module. The Dialog Manager encodes the ``state'' of the dialog as well, which gets passed to the Natural Language Generation module, which formulates the system's response to the user.}
\label{fig:dialog-system}
\end{figure*}

\section{Task-Oriented Dialog Systems, Intent Classification, and Slot-Filling}\label{sec:dialog-systems}

In this section, we briefly introduce task-oriented dialog systems, intent classification, and slot-filling.
We then discuss various utterance types, data sources for creating intent classification and slot-filling datasets, and the metrics commonly used to evaluate models on them.

\subsection{Task-Oriented Dialog Systems}
Task-oriented dialog systems, along with the essential components of intent classification and slot-filling, are particularly relevant in recent years because of the growth of applications for goal-driven dialog systems in areas and industries such as banking and personal finance, e-commerce, technical support, healthcare, travel, etc. \citep{wen-etal-2019-data-emnlp-lecture-tutorial-polyai}.
Typical architectures for task-oriented dialog systems incorporate several modules, including automatic speech recognition (ASR), natural language understanding (NLU), business logic, dialog management, and natural language generation (NLG) modules. 
Figure~\ref{fig:dialog-system} illustrates how these components interact. Within the natural language understanding module, text is classified into (an optional) domain as well as a finer-grained intent. 
This intent frequently maps to a core functionality that the dialog system supports. For example, the user query \emph{``find me flights going to Orlando from Detroit on Friday''} could map to a system-supported intent class called \texttt{find\_flights}. This \texttt{find\_flights} intent class might also fall within the \texttt{travel} domain.
\footnote{In this example, we say that the \texttt{find\_flights} intent \emph{might} fall within the \texttt{travel} domain because different system designers have different requirements---in principle, a system designer could have any intent map to any domain, just as any user query could fall within any intent.}

Identifying the intent of the user's query is but one of the tasks of the NLU module. An additional step is to extract the important entities from the query, which can then be used as inputs to the subsequent modules. 
For instance, in our example \emph{``find me flights going to Orlando from Detroit on Friday''}, the important entities are the city names ``Orlando'' and ``Detroit'', as well as the day ``Friday''. We call these entities \emph{slots}. 
In our hypothetical dialog system, these extracted slots might map to the \emph{slot types} of \texttt{destination\_location}, \texttt{departure\_location}, and \texttt{departure\_day}. More specifically, ``Orlando'' is the extracted \emph{slot value} for the \texttt{destination\_location} \emph{slot type}.

The task of extracting these slots is known interchangeably as \emph{slot-filling}, \emph{slot/entity extraction}, or \emph{entity/slot tagging}. 
Simple pattern matching systems that look for city names and days of the week might be a straightforward initial solution to the slot-filling task, but such an approach has downsides: First, human users might refer to the departure and destination locations by airport names (like ``DTW'' and ``Love Field''), in which case the list of possible acceptable location names could become very large. 
Second, the contextual information surrounding the slot values helps indicate the appropriate slot types. 
In our example, the contextual span ``going to'' indicates that ``Orlando'' is the destination, while ``from'' indicates that ``Detroit'' is where the user seeks to depart.
Standard named entity recognition (NER) models pre-trained to extract locations and dates might also not be appropriate here, as such systems may not be trained to distinguish between a departure and destination location, even though these two locations might be named-entities.
Instead, typical machine learning-driven slot extraction modules often rely on sequential models such as stochastic finite state transducers, sequential classifiers, conditional random fields, recurrent neural networks (RNNs) (e.g., \citealp{yao2013recurrent, mesnil2013investigation, rnn-journal}), LSTMs (e.g., \citealp{lstm-slt}), and transformers and attention-based models (e.g., \citealp{devlin-etal-2019-bert, wu-etal-2020-tod-bert, wu-etal-2020-slotrefine}).

The extracted slots, along with the query's predicted intent, are then consumed by the dialog manager, whose role is to determine and construct the system's response to the user's query. The dialog manager may in turn call on \emph{business logic}: the application-specific layer that turns a predicted intent and its filled slots into an action, by querying knowledge bases or APIs and applying the rules of the particular application. In our running example, it could query a flight booking database.
Business logic is also where slot values are resolved to canonical entities---knowing that ``DTW'' and ``Detroit'' both denote Detroit Metropolitan Wayne County Airport lets the system find flights from that specific airport.
The dialog manager uses output from both the NLU and business logic modules, as well as knowledge of the conversation history, to produce information that is used by the natural language generation (NLG) module to construct a response to the user. In typical systems, the NLG module may respond to the user using text or text-to-speech with speech synthesis.

\subsection{Utterance Types}

\begin{figure*}
\centering
\resizebox{0.85\textwidth}{!}{%
\begin{tikzpicture}[
  font=\small,
  sys/.style={draw=black!75, fill=figSys, align=left, text width=4.3cm,
              inner sep=2.2mm, font=\small},
  usr/.style={draw=black!75, fill=figUsr, align=left, text width=4.3cm,
              inner sep=2.2mm, font=\small\itshape},
  tag/.style={font=\bfseries},
]
\def\ind{1.3}   

\node[sys] (s0) at (0,0) {Hello, how may I help you?};
\node[usr, below=3.5mm of s0.south west, anchor=north west, xshift=\ind cm] (u1)
  {Can you tell me the balance of my checking account?};
\node[sys, below=3.5mm of u1.south west, anchor=north west, xshift=-\ind cm] (s1)
  {Sure, your Checking Account balance is \$2,529.18.};
\node[usr, below=3.5mm of s1.south west, anchor=north west, xshift=\ind cm] (u2)
  {And what about my savings?};
\node[sys, below=3.5mm of u2.south west, anchor=north west, xshift=-\ind cm] (s2)
  {Your Savings Account balance is \$12,849.91.};
\node[usr, below=3.5mm of s2.south west, anchor=north west, xshift=\ind cm] (u3)
  {Did the Tigers win last night?};
\node[sys, below=3.5mm of u3.south west, anchor=north west, xshift=-\ind cm] (s3)
  {Sorry, I don't know how to answer that.};

\coordinate (tagL) at (-2.6, 0);
\node[tag, anchor=east] at (tagL |- s0.west) {S0)};
\node[tag, anchor=east] at (tagL |- u1.west) {U1)};
\node[tag, anchor=east] at (tagL |- s1.west) {S1)};
\node[tag, anchor=east] at (tagL |- u2.west) {U2)};
\node[tag, anchor=east] at (tagL |- s2.west) {S2)};
\node[tag, anchor=east] at (tagL |- u3.west) {U3)};
\node[tag, anchor=east] at (tagL |- s3.west) {S3)};

\node[usr] (u4) at (9.6, 0) {Tell me my most recent transaction.};
\node[sys, below=3.5mm of u4.south west, anchor=north west, xshift=-\ind cm] (s4)
  {Your last transaction was today for \$62.88 at IKEA.};
\node[usr, below=3.5mm of s4.south west, anchor=north west, xshift=\ind cm] (u5)
  {What was my largest transaction this week and what is my withdrawal limit?};
\node[sys, below=3.5mm of u5.south west, anchor=north west, xshift=-\ind cm] (s5)
  {Your largest transaction this week was on Saturday for \$102.84 at Home Depot. Your withdrawal limit is \$1000.};
\node[usr, below=3.5mm of s5.south west, anchor=north west, xshift=\ind cm] (u6)
  {Thank you!};
\node[sys, below=3.5mm of u6.south west, anchor=north west, xshift=-\ind cm] (s6)
  {I'm happy to help!};

\coordinate (tagR) at (5.7, 0);
\node[tag, anchor=east] at (tagR |- u4.west) {U4)};
\node[tag, anchor=east] at (tagR |- s4.west) {S4)};
\node[tag, anchor=east] at (tagR |- u5.west) {U5)};
\node[tag, anchor=east] at (tagR |- s5.west) {S5)};
\node[tag, anchor=east] at (tagR |- u6.west) {U6)};
\node[tag, anchor=east] at (tagR |- s6.west) {S6)};
\end{tikzpicture}}
\caption{Example dialog between a user (\textbf{U}) and a dialog system (\textbf{S}). Utterance \textbf{U2} is an example of a follow-up query, one that implies information contained in the previous dialog turn. Utterance \textbf{U3} is an example of an out-of-scope query. Utterance \textbf{U5} is an example of a simple multi-intent query where two queries (\emph{``What was my largest transaction this week''} and \emph{``what is my withdrawal limit''}) are joined by a simple conjunction.}
\label{fig:example-dialog}
\end{figure*}
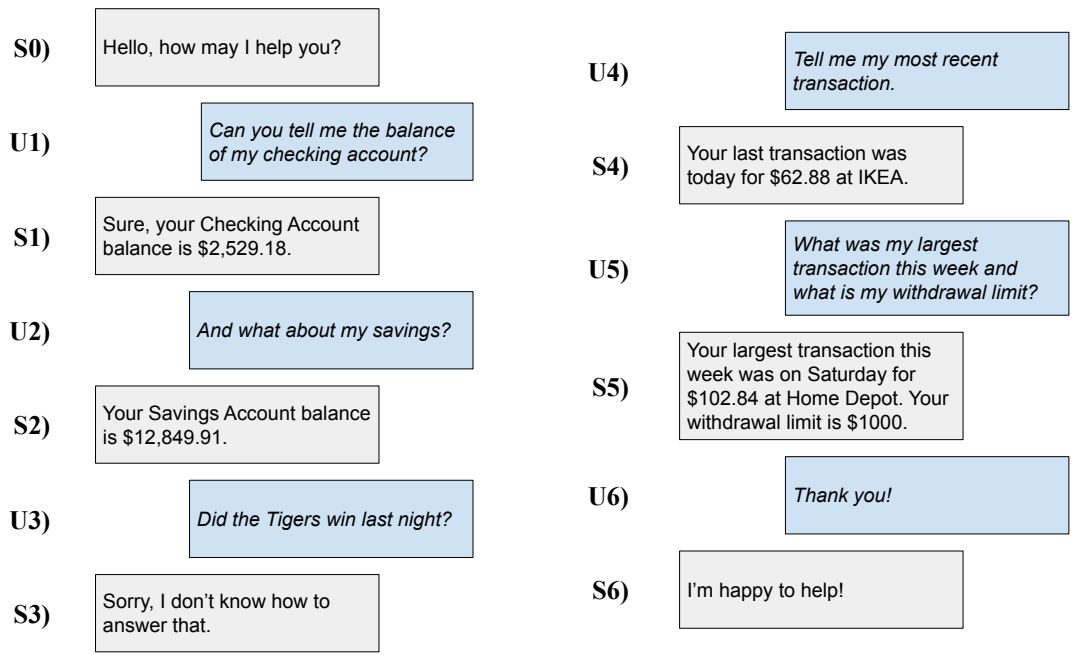

While not exhaustive, Figure~\ref{fig:example-dialog} shows examples of several types of utterances, knowledge of which will be helpful for the rest of this survey.
For the example in Figure~\ref{fig:example-dialog}, we pretend that the task-oriented dialog system is knowledgeable of the user's bank account.
\textbf{U1} is the initial, or ``root'', user query or utterance.
The initial utterance is \emph{in-scope} with respect to the dialog system, and (assuming there is a corresponding intent for balance queries and that the intent classifier correctly identified utterance \textbf{U1} as a balance query) the system responds accordingly.
\textbf{U2} is a \emph{follow-up} utterance; one that depends on the previous user query and/or system response.
Most of the datasets surveyed in this paper do include initial/root queries, but few have follow-up utterances.
The user's \textbf{U3} utterance is out-of-scope (or out-of-domain) with respect to the dialog system;
assuming the system correctly flags \textbf{U3} as such, it might respond with an appropriate fallback message in \textbf{S3}.
The next utterance worthy of note in this example is \textbf{U5}, which consists of two queries: \emph{what was my largest transaction this week}; and \emph{what is my withdrawal limit}.
This is a \emph{multi-intent} utterance.
We will see that few of the datasets surveyed in this paper contain multi-intent utterances.

\par\vspace{\intextsep}\noindent\begin{minipage}{\columnwidth}\centering
    \centering
    \adjustbox{max width=\columnwidth}{\fbox{\begin{minipage}{0.99\columnwidth}
    \small
    \begin{center}
        \textbf{Rephrase an original question or statement}
    \end{center}
    \footnotesize
    Suppose you have an intelligent device such as Amazon Alexa, Apple Siri, or Google Assistant.\\[3pt]
    Given an original phrase, provide 5 different ways of saying the same phrase.\\[3pt]
    \textbf{Original phrase: {\color{blue}``what is the weather like in Cleveland''}}
    \end{minipage}}}
    \vspace{1mm}
    
    \adjustbox{max width=\columnwidth}{\fbox{\begin{minipage}{0.99\columnwidth}
    \small
    \begin{center}
        \textbf{Rephrase an original question or statement}
    \end{center}
    \footnotesize
    Suppose you have an intelligent device such as Amazon Alexa, Apple Siri, or Google Assistant.\\[3pt]
    Given an original phrase, provide 5 different ways of saying the same phrase.\\[3pt]
    \textbf{Original phrase: {\color{blue}``what is the weather like in Cleveland''}}\\[3pt]
    \textbf{Don't use the words ``{\color{red}weather}'' or ``{\color{red}what}'' in your responses.}
    \end{minipage}}}
    \vspace{1mm}
    
    \adjustbox{max width=\columnwidth}{\fbox{\begin{minipage}{0.99\columnwidth}
    \small
    \begin{center}
        \textbf{Scenario:}
    \end{center}
    \footnotesize
    Determine the type of aircraft used on a flight from Cleveland to Dallas that leaves before noon.
    \end{minipage}}}
    \vspace{1mm}

    
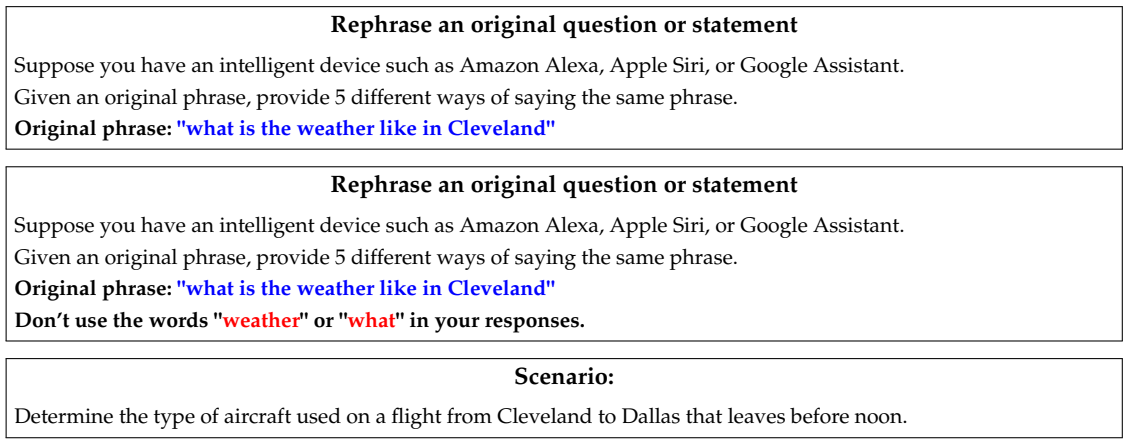
\captionof{figure}{\label{fig:prompts}
    Examples of crowdsource prompts for data collection.
    The top two prompts ask crowd workers to paraphrase utterances, and may optionally require or restrict certain lexical features in the paraphrases (prompts adapted from \citet{larson-etal-2020-iterative}).
    The bottom prompt is an example of a scenario prompt from \citet{dahl-etal-1994-expanding-atis}.
    \vspace{-3mm}}
\end{minipage}\par\vspace{\intextsep}

\subsection{Sources of Data}
As we will see in our survey of various datasets, there are several common ways in which dataset creators collect, gather, or otherwise produce data for their datasets.
The survey by \citet{mohammad-survey} provides a typology of data source type, which we adopt and adapt here.
This typology includes crowdsourcing, wherein crowd workers (e.g., from Amazon Mechanical Turk) are prompted by the dataset designers to respond to certain scenarios or produce paraphrases of example utterances directed toward certain intent categories with optional slots.
Figure~\ref{fig:prompts} lists three crowdsourcing prompt examples.
We also note that in some cases, the designers of a dataset (or others knowledgeable in the creation of the dataset) may serve as providers of utterances; additionally, experts are often used in translating data from one language to another.

\citet{mohammad-survey} also observed that utterances may be generated from templates, an approach used to produce the \emph{Leyzer} dataset \citep{leyzer}, for instance.
An example template used to construct a subset of utterances in the \emph{Leyzer} dataset is shown in Figure~\ref{fig:grammar-example}.
Datasets may also be built from existing utterances that were provided to real dialog systems.
Finally, we observe that many of the datasets discussed below are derived in some way from prior datasets.
This includes datasets that were translated from one dialog dataset's language to a target language (e.g., \emph{Multilingual ATIS} \citep{multilingual-atis} and \emph{MultiATIS++} \citep{multiatispp}), or combined from several other prior datasets (e.g., \emph{xSID} \citep{van-der-goot-etal-2021-masked-xSID} and \emph{Redwood} \citep{Larson2022-redwood}).
Below, we refer to these data sources as \emph{crowd}, \emph{experts}, \emph{generated}, \emph{users}, and \emph{derived}.
\par\vspace{\intextsep}\noindent\begin{minipage}{\columnwidth}\centering
    \includegraphics[width=\columnwidth]{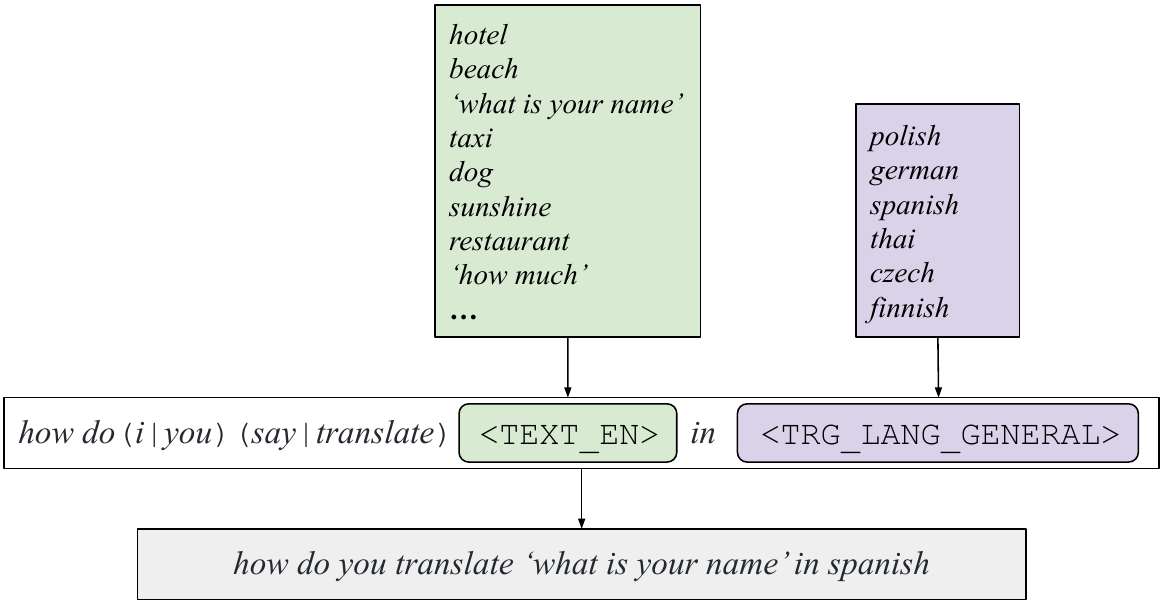}
    \captionof{figure}{Example grammar for generating utterances for a \texttt{translate} intent (adapted from \citet{leyzer}).}
    \label{fig:grammar-example}
\end{minipage}\par\vspace{\intextsep}




\subsection{Common Evaluation Metrics}\label{sec:evaluation-metrics}

We will occasionally refer to model performance on certain datasets, so this section defines some commonly used evaluation metrics for evaluating slot-filling and intent classification models.

\paragraph{\textbf{Intent Classification Evaluation Metrics}}~ Determining the intent of an utterance is typically framed as a classification problem, and is evaluated with the standard classification metrics.
\emph{Accuracy}---the fraction of test utterances whose intent is predicted correctly---is the most commonly reported.
\emph{Precision}, $tp/(tp+fp)$, and \emph{recall}, $tp/(tp+fn)$, differ from it only in their denominators, measuring correct predictions against those the model made and against those it should have made, respectively; their harmonic mean is the \emph{F1} score.
F1 is often preferred over accuracy on datasets with large class imbalances, a property we quantify for the corpora surveyed here in Section~\ref{sec:discussion-future-work}.

In the multi-class settings typical of intent classification, precision and recall may be \emph{macro}-averaged, computing the metric separately for each of the $|C|$ intent classes and then averaging over classes, or \emph{micro}-averaged, pooling true and false positives across all classes before computing it.
The distinction bears on comparability: macro-averaging weights a rare intent as heavily as a frequent one, so the two can diverge substantially on the imbalanced corpora we discuss below, and reported results do not always state which was used.

\paragraph{\textbf{Slot-Filling Evaluation Metrics}}~ Most datasets treat slot-filling similarly to that of Named Entity Recognition (NER), and data labels are represented as \emph{Begin}, \emph{Inside}, and \emph{Outside} (or \emph{B-I-O}) tags.
Table~\ref{tab:bio_tags} shows \emph{B/I/O} tags on an example utterance.
Most evaluations on slot-filling datasets measure the F1 score, where each slot type (e.g., origin, destination, and the null O type\footnote{Often, the O label forms the overwhelming majority of label tags in slot-filling datasets.}) is considered as a classification label.

\begin{table}[!htbp]
    \centering
    \caption{B/I/O label representation for slot-filling data.}
    \adjustbox{max width=\columnwidth}{
    \begin{tabular}{ccccccccc}
    \toprule
        \emph{find} & \emph{me} & \emph{a} & \emph{flight} & \emph{from} & \emph{San} & \emph{Jose} & \emph{to} & \emph{Nashville}\\
    \midrule
        O & O & O & O & O & B-\texttt{origin} & I-\texttt{origin} & O & B-\texttt{destination} \\
    \bottomrule
    \end{tabular}}
    \label{tab:bio_tags}
\end{table}

\paragraph{\textbf{Joint Model Evaluation Metrics}}~ Evaluations of joint models may use individual metrics for both intent classification and slot-filling tasks.
Additionally, \emph{exact match accuracy} is often used as a combined measure of intent classification and slot-filling performance.
In exact match accuracy, the numerator of the accuracy ratio is the number of utterances whose predicted labels (including intent and all slot predictions) completely match the ground-truth labels.


\section{Joint Intent Classification and Slot-Filling Datasets}\label{sec:joint-datasets}

We begin our survey of datasets by cataloging those for the task of joint intent classification and slot-filling.
These datasets have intent and slot annotations; while they are suitable for joint modeling tasks, they can also be used as benchmarks for intent classifiers and slot-filling models individually.
Datasets surveyed in this section are summarized in Table~\ref{tab:joint-datasets} (English datasets) and Table~\ref{tab:other-datasets} (non-English datasets).

Throughout this section and Section~\ref{sec:intent-classification-datasets}, we supplement our qualitative descriptions with several quantitative measures that we compute uniformly across datasets so that they are comparable.
We summarize lexical variety using vocabulary size and type-token ratio, and intent-distribution balance using the normalized Shannon entropy of the intent labels (with 1.0 denoting a perfectly uniform distribution).
As a proxy for intent separability and task difficulty, we report the accuracy of a five-fold cross-validated linear (logistic-regression) probe trained on fixed, off-the-shelf multilingual sentence embeddings.\footnote{We encode utterances with the \texttt{paraphrase-multilingual-MiniLM-L12-v2} sentence-transformer model \citep{reimers-2019-sentence-bert}. Because the encoder and probe are held fixed across all datasets, these accuracies are intended as \emph{comparative} diagnostics rather than state-of-the-art results, which are typically obtained by fine-tuning much larger models.}

We quantify train/test redundancy in two ways: lexically, as the \emph{out-of-vocabulary (OOV) rate}---the fraction of test tokens (and of test word types) unseen in the training split---and semantically, as the fraction of test utterances whose nearest training utterance exceeds a cosine similarity of 0.90 in the same embedding space.
The former is encoder-free and therefore does not inherit the biases of a particular pretrained model; the latter captures paraphrase-level redundancy that lexical matching misses.\footnote{All lexical measures use a single Unicode-aware tokenizer applied uniformly: scripts written with word delimiters are segmented into words, while Chinese, Japanese, and Thai---which lack them---are segmented into characters, avoiding any dependence on a language-specific word segmenter. Character-level counts are not commensurable with word-level ones, so vocabulary and OOV figures for those languages should be compared only among themselves.}
We also measure redundancy \emph{between} datasets, as the Jaccard similarity of two training vocabularies together with the directional coverage of each vocabulary by the other (the fraction of dataset $A$'s word types that also occur in $B$), which distinguishes genuine overlap from mere containment of a small corpus in a large one.
Finally, we estimate label quality with two complementary signals: confident learning \citep{northcutt-2021-confident}, applied to the probe's out-of-sample class probabilities to flag likely mislabeled examples, and a semantic-consistency check that surfaces near-duplicate utterances (cosine similarity above 0.95) bearing different intent labels.
These probe-based and label-quality measures require single-intent labels, and so we omit them for multi-intent and slot-only datasets.


\paragraph{\textbf{ATIS}}~ The \emph{Air Travel Information System} (\emph{ATIS}) corpus is by far the oldest dataset for evaluating intent classification and slot-filling models. 
It was assembled from a series of collections carried out in the early 1990s, beginning with \emph{ATIS-0} \citep{hemphill-etal-1990-atis}.
That first round consists of transcribed recordings of Texas Instruments employees querying what they took to be a flight scheduling system.
The system was in fact a ``wizard''---two experts who transcribed each spoken query and translated it into a query against the Official Airline Guide database.
\emph{ATIS-0} is thus an early instance of task-oriented \emph{wizard-of-oz} collection, in which one party pursues a goal while the other imitates a system's responses.
Subsequent rounds by \citet{hirschman-1992-multi-location-atis}, \citet{hirschman-etal-1993-multi-lication-atis}, and \citet{dahl-etal-1994-expanding-atis} produced \emph{ATIS-1}, \emph{ATIS-2}, and \emph{ATIS-3}; utterances drawn from these form the \emph{ATIS} benchmark researchers use today, which has
22 intent categories and 79 slot types.\footnote{The precise intent and slot counts reported for \emph{ATIS} vary across its many redistributions and preprocessing choices---a lack of standardization that is itself a recurring concern for the benchmark, with published figures ranging from 18 to 26 intents and from 79 to 84 slot types. The counts we report are measured directly on the copy we analyze, the widely used 4,978-utterance training split: it contains 17 base intent categories, or 22 once multi-intent utterances (whose labels combine several intents, e.g., \texttt{airfare+flight}) are counted as distinct classes, and 79 slot types.}
Table~\ref{tab:atis-examples} shows examples; the benchmark has also been translated into several other languages, which we discuss below.

\begin{table}[tbp]
    \centering
    \caption{Example utterances with slots from the \emph{ATIS} corpus.}
    \scalebox{0.670}{
    \begin{tabular}{cl}
    \toprule
        \textbf{Intent} & \textbf{Utterance}  \\
        \midrule
        \texttt{airfare}         & $\text{show me the} \atisCostRelative{cheapest} \text{~~fare from} \atisFromLoc{dallas} \text{to~~} \atisToLoc{baltimore}$\\
        \texttt{flight}          & $\text{list the takeoffs and landings at~} \atisAirport{general mitchell international}$\\
        \texttt{airline}         & $\text{show~ me~ the~ airlines~ that~ fly~ from} \atisFromLoc{denver} \text{to~~~~~} \atisToLoc{san francisco}$\\
        \texttt{flight}          & $\text{flights from ~}\atisFromLoc{baltimore}\text{to~~~} \atisToLoc{philadelphia}\text{~ please}$\\
        \texttt{abbreviation}    & $\text{what~ is} \atisAirportCode{ewr}$\\
        \texttt{ground\_service} & $\text{is there ground transportation from the} \atisAirport{milwaukee airport} \text{to downtown}$\\
        \texttt{aircraft}        & $\text{list aircraft types that fly between} \atisFromLoc{boston} \text{ and} \atisToLoc{san francisco}$\\
        \texttt{quantity}        & $\text{how many airports does} \atisCityName{oakland} \text{have}$\\
        \texttt{flight}          & $\text{show ~me ~all ~flights ~to~~~} \atisToLoc{philadelphia} \text{~~~~in the} \atisDepartTimePeriodOfDay{evening}$\\
        \texttt{airline}         & $\text{what~ is~ airline} \atisAirlineCode{us}$\\
        \texttt{flight}          & $\text{find~ me~ a~ flight~ leaving} \atisFromLoc{boston} \text{at~~~} \atisDepartTime{12 o'clock}$\\
        \bottomrule
    \end{tabular}}
    \label{tab:atis-examples}
\end{table}

\emph{ATIS} is so nearly synonymous with the paradigm of Section~\ref{sec:dialog-systems} that \citet{niu-penn-2019-rationally-reappraising-atis} simply call such systems ``\emph{ATIS}-based,'' and \citet{Casanueva2022-nlu} remark that ``remarkably, \emph{ATIS} is still considered at present as one of the main go-to datasets in NLU research.''
Its longevity has nonetheless drawn sustained criticism, which our own measurements bear out.
\citet{atis-shallow} argue that \emph{ATIS} is too ``shallow'' to challenge contemporary models and that many residual errors reflect annotation mistakes or genuine ambiguity, while \citet{niu-penn-2019-rationally-reappraising-atis} fault its ``small amount of training data and an overall lack of lexical and syntactic variety,'' and find a rule-based parser competitive with neural models once annotation errors are corrected.
That narrowness is pronounced: the training set contains only 886 distinct word types (a type-token ratio of 0.016), roughly one-sixth as lexically varied, on a length-normalized basis, as a comparable joint dataset such as \emph{Snips} (11,567 word types).
\citet{larson-etal-2020-data} further observe that the \texttt{from\_loc.city\_name} and \texttt{to\_loc.city\_name} slots overwhelmingly follow the tokens ``from'' and ``to''---a trigger-word regularity we take up in Section~\ref{subsec:slot-schemes}.

Three further properties compound the problem.
\emph{ATIS} is severely imbalanced: the \texttt{flight} intent accounts for the large majority of utterances, yielding a normalized intent entropy of just 0.37 against values near 1.0 for \emph{Snips} and \emph{Clinc-150}, so a classifier can score well largely by favoring the majority class---and its intents are correspondingly poorly separated in embedding space (a slightly negative mean silhouette coefficient).
Its test set is highly redundant with its training set: 43\% of test utterances have a near-duplicate (embedding cosine similarity above 0.90) in training, against 36\% for \emph{Clinc-150} and 17\% for \emph{Snips}.
And consistent with the annotation-error concerns above, near-identical utterances carry inconsistent labels---\emph{``give me the flights and fares on december twenty seventh from indianapolis''} is labeled \texttt{airfare+flight}, while \emph{``give me the flights with the fares on december twenty seventh $\ldots$''} is labeled \texttt{flight} alone.
Little wonder that the joint models surveyed by \citet{weld2021survey} have approached ceiling on \emph{ATIS}; it remains a cornerstone of NLU evaluation all the same.




\paragraph{\textbf{Datasets derived from ATIS}}~ Most corpora for intent classification and slot-filling are English, which makes it harder to build and evaluate dialog systems for other languages; \emph{ATIS} has accordingly been translated more often than any other corpus in this survey.
\textbf{\emph{Multilingual ATIS}} \citep{multilingual-atis} renders a subset of the English corpus in Turkish and Hindi, translated manually by native speakers with slots re-annotated in the target language by crowdsourcing, and \citet{susanto-lu-2017-neural} likewise produced an Indonesian version.
\textbf{\emph{MultiATIS++}} \citep{multiatispp} extends the family to nine languages across four families---English, Spanish, Portuguese, French, German, Chinese, Japanese, Hindi, and Turkish---using professional native translators who also annotated the target-language slots; it retains the 18 intents and 84 slot types of its English partition, though the smaller Hindi and Turkish portions cover only a subset of them.
\textbf{\emph{PhoATIS}} \citep{JointIDSF-phoatis}, with 28 intent labels and 82 slot types, was the first publicly available intent detection and slot-filling dataset for Vietnamese.
\textbf{\emph{Persian ATIS}} \citep{akbari-etal-2023-persian-atis} covers a language \emph{MultiATIS++} omits, and was produced differently again: machine translation, then verification and correction by native speakers, with slots re-annotated afterwards; it retains all 5,871 utterances of the original (4,978 training and 893 evaluation).
The family therefore spans three translation regimes---manual, professional, and machine translation with human post-editing---a distinction worth bearing in mind when comparing results across its members, since machine- and crowd-translated partitions carry measurably higher estimated label noise than hand-curated corpora (Section~\ref{sec:discussion-future-work}).

\paragraph{\textbf{Braun Collection}}~ Three datasets comprise the \emph{Braun Collection} \citep{braun-etal-2017-evaluating}: the \textbf{\emph{Chatbot Corpus}}, the \textbf{\emph{Web Applications}} dataset, and the \textbf{\emph{ask ubuntu}} dataset (the latter two being part of what \citet{braun-etal-2017-evaluating} call the \emph{StackExchange Corpus}).
Both the \emph{Web Applications} and \emph{ask ubuntu} datasets were constructed by scraping queries from various StackExchange forums like \texttt{askubuntu.com} and \texttt{webapps.stackexchange.com}.
As such, the utterances in these two datasets were not originally intended for dialog systems, but they are nonetheless dialog-style utterances directed at various intents related to software support.
Both the \emph{Web Applications} and \emph{ask ubuntu}
datasets consist of English utterances.
The \emph{Chatbot Corpus} is composed of real user utterances from a dialog system for public transit queries in Munich, Germany.
While the utterances are in English, they do contain many German place names.

The number of intents for each dataset in the \emph{Braun Collection} is low, ranging from 2 to 8.
The number of training samples is also quite low, with the largest being the \emph{Chatbot Corpus} at 206 utterances (100 of them training), and the smallest the \emph{Web Applications} dataset at 89 (30 training).
(This is compared to the thousands of utterances in the other datasets listed in Table~\ref{tab:joint-datasets}).
As such, the \emph{Web Applications} and \emph{ask ubuntu} datasets offer extreme training scenarios, with \emph{Web Applications} containing no intent with more than 7 training samples, and \emph{ask ubuntu} having a maximum of 17 training samples per intent.
The number of slot types is also low for each of the datasets, ranging from 3 to 5. 

\begin{table*}[tbp]
    \centering
    \caption{Joint intent classification and slot-filling datasets with English language utterances. (Note that we present statistics for the English versions of \emph{Leyzer}, \emph{Facebook}, \emph{MTOP}, and \emph{xSID}.)}
    \scalebox{0.75}{ 
    \begin{tabular}{lccccc}
    \toprule
        \textbf{Dataset} & \textbf{Intents} & \textbf{Slots} & \textbf{\# Utterances} & \textbf{Source} & \textbf{License} \\
        \midrule
        \emph{ATIS} \citep{hemphill-etal-1990-atis, hirschman-1992-multi-location-atis} & 22 & 79 & 5,871 & Crowd & \textsc{ldc} \\ 
        \citep{hirschman-etal-1993-multi-lication-atis, dahl-etal-1994-expanding-atis} & & & & & \\
        \emph{ask ubuntu} \citep{braun-etal-2017-evaluating} & 5 & 3 & 162 & Users & \textsc{cc-by-sa 3.0}\\
        \emph{Chatbot} \citep{braun-etal-2017-evaluating}  & 2 & 5 & 206 & Users & \textsc{cc-by-sa 3.0}\\
        \emph{Web Applications} \citep{braun-etal-2017-evaluating}  & 8 & 3 & 89 & Users & \textsc{cc-by-sa 3.0}\\
        \emph{Snips} \citep{snips} & 7 & 72 & 14,484 & Crowd & \textsc{cc0 1.0} \\
        \emph{TOP} \citep{gupta-etal-2018-semantic}  & 25 & 36 & 44,783 & Crowd & --- \\
        \emph{HWU-64} \citep{XLiu.etal:IWSDS2019-hwu} & 64 & 54 & 25,716 & Crowd & \textsc{cc-by-sa 3.0}\\
        \emph{Facebook} \citep{schuster-etal-2019-cross-lingual} & 12 & 11 & 43,323 & Crowd & \textsc{cc-by-sa} \\
        \emph{TOPv2}  \citep{chen-etal-2020-low-topv2} & 80 & 82 & 181,000 & Crowd & --- \\
        \emph{Leyzer} \citep{leyzer}  & 186 & 86 & 3,892 & Generated & \textsc{cc-by-nc-nd 4.0}\\
        \emph{MixATIS} \citep{qin-etal-2020-agif-mixatis-mixsnips} & 24 & 83 & 20,000 & Derived & ---\\
        \emph{MixSNIPS} \citep{qin-etal-2020-agif-mixatis-mixsnips} & 7 & 72 & 50,000 & Derived & ---\\
        \emph{CSTOP} \citep{cstop} & 19 & 10 & 5,803 & Expert & ---\\ 
        \emph{MTOP} \citep{li2021mtop} & 117 & 78 & 22,288 & Crowd & --- \\
        \emph{xSID} \citep{van-der-goot-etal-2021-masked-xSID} & 16 & 41 & 44,405 & Derived & \textsc{cc-by-sa 4.0} \\
        \emph{NLU++} \citep{Casanueva2022-nlu} & 62 & 17 & 3,080 & Users & --- \\
        \emph{MASSIVE} \citep{fitzgerald-etal-2023-massive} & 60 & 55 & 1,000,000 & Translated & \textsc{cc-by 4.0} \\
        \emph{BlendX} \citep{yoon-etal-2024-blendx} & --- & --- & --- & Derived, Generated & --- \\
        \bottomrule
    \end{tabular}}
    \label{tab:joint-datasets}
\end{table*}

\paragraph{\textbf{Snips}}~ Like \emph{ATIS}, the \emph{Snips} corpus \citep{snips} is a commonly-used benchmark for both intent classification and slot-filling (and their joint) tasks.
Its intents cover the kinds of requests a user might make of a general-purpose virtual assistant, such as asking for the weather or playing music.
Data for \emph{Snips} was collected using crowdsourcing, in which crowdworkers were asked to respond to scenario-like prompts targeting particular intent categories with certain slot values.
For instance, in the example given in \citet{snips}, workers were asked to respond to the following prompt:
\begin{center}
    \texttt{Intent: The user wants to switch the lights on; slot: (bedroom)}
\end{center}
to which the worker might provide the utterance:
\begin{center}
    \emph{I want the lights in the bedroom on right now}
\end{center}
The composition of the \emph{Snips} corpus is straightforward: 
In total, \emph{Snips} has 72 slot types---a subset of which appear across multiple intents---and there are 7 intent categories.
Each intent is therefore backed by some 2,000 training samples on average, a generous allowance by the standards of this survey.
Despite having only seven intents, \emph{Snips} is lexically far richer than intent classification datasets with many more classes: its training vocabulary of over 11,500 word types exceeds that of \emph{Clinc-150} (5,503 types across 150 intents) and \emph{Banking-77} (2,384 types across 77 intents).
This richness stems from the entity-heavy nature of the \emph{Snips} intents (e.g., \texttt{PlayMusic}, \texttt{SearchCreativeWork}, and \texttt{BookRestaurant}), whose utterances are populated with song, artist, movie, and place names.

Like \emph{ATIS}, \emph{Snips} is a well-known corpus for benchmarking joint models, and like \emph{ATIS}, joint models have been approaching near-perfection on \emph{Snips} \citep{weld2021survey}.
Notably, this saturation holds even though \emph{Snips} is among the least redundant of these benchmarks: only 17\% of its test utterances have a near-duplicate (embedding cosine similarity above 0.90) in the training set, compared to 36\% for \emph{Clinc-150} and 67\% for \emph{Banking-77}.
A simple linear probe over sentence embeddings already classifies its seven intents at 96.9\% accuracy, with the few remaining errors falling among genuinely ambiguous ``search'' intents---for instance, \texttt{SearchScreeningEvent} versus \texttt{SearchCreativeWork}.

\paragraph{\textbf{Almawave-SLU}}~ The \emph{Almawave-SLU} dataset \citep{almawave} is similar to \emph{Multilingual ATIS} insofar as it is translated from an existing English dataset, but instead of using ATIS as the source dataset, \emph{Almawave-SLU} is derived from \emph{Snips}.
\emph{Almawave-SLU} consists of 7,142 samples translated from the English \emph{Snips} dataset into Italian.
Whereas \emph{Multilingual ATIS} (as well as several \emph{ATIS}-derived multilingual datasets discussed above) were translated manually, samples in \emph{Almawave-SLU} were first translated using machine translation, then verified and---if needed---corrected by human annotators. 

\paragraph{\textbf{MixATIS and MixSNIPS}}~
Noticing a relative lack of multi-intent datasets for evaluating intent classification and slot-filling models, \citet{qin-etal-2020-agif-mixatis-mixsnips} artificially create multi-intent queries by joining queries from single-intent datasets.
They create \emph{MixSNIPS} and \emph{MixATIS} by joining queries from the \emph{Snips} and \emph{ATIS} datasets, respectively.
Multi-intent queries are constructed by joining queries with conjunctions like ``and'', ``,'' (comma), ``and also'', ``and then'', etc. Both \emph{MixATIS} and \emph{MixSNIPS} consist of queries having between 1 and 3 intents (at a ratio of 0.3:0.5:0.2).
Figure~\ref{fig:mix-snips-example} shows an example of two single-intent utterances joined with a conjunction.
While \emph{MixSNIPS} and \emph{MixATIS} are notable because they are dedicated multi-intent datasets, they are both limited by their parent corpora (\emph{ATIS} and \emph{Snips}) as well as the relatively few conjunctions that are used to connect queries.
\par\vspace{\intextsep}\noindent\begin{minipage}{\columnwidth}\centering
    \includegraphics[width=\columnwidth]{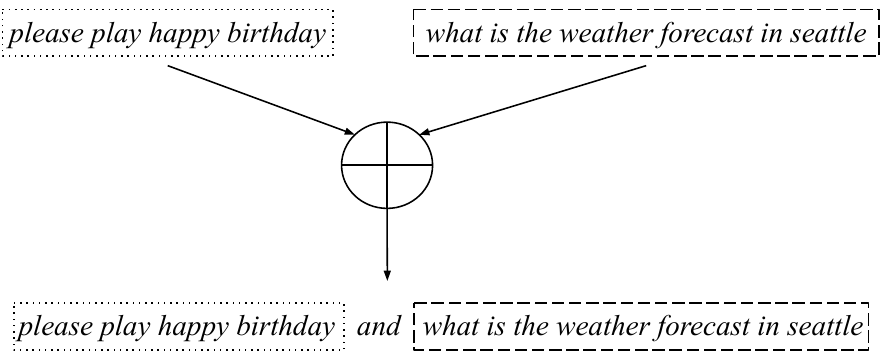}
    \captionof{figure}{Multi-intent samples from \emph{MixSNIPS} and \emph{MixATIS} are formed by joining utterances together with conjunctions like ``and''.}
    \label{fig:mix-snips-example}
\end{minipage}\par\vspace{\intextsep}

\paragraph{\textbf{BlendX}}~
\citet{yoon-etal-2024-blendx} extend the approach of \emph{MixATIS} and \emph{MixSNIPS} to create \emph{BlendX}, a suite of multi-intent datasets with more complex and varied utterance patterns.
Where \emph{MixATIS} and \emph{MixSNIPS} join single-intent queries using a small fixed set of conjunctions, \emph{BlendX} employs a combination of rule-based heuristics and ChatGPT-assisted generation with a similarity-driven utterance selection strategy to produce blended multi-intent utterances that are more linguistically diverse.
\emph{BlendX} is derived from four source datasets---\emph{ATIS}, \emph{Banking-77}, \emph{Clinc-150}, and \emph{Snips}---yielding four corresponding \emph{BlendX} variants, and the authors also introduce three new metrics for assessing utterance complexity based on word count, conjunction use, and pronoun usage.
Two of those four sources, \emph{Banking-77} and \emph{Clinc-150}, are intent classification corpora carrying no slot annotations (both are discussed in Section~\ref{sec:intent-classification-datasets}), so only the \emph{ATIS}- and \emph{Snips}-derived variants support the joint task; we catalog \emph{BlendX} here on the strength of those two.
The difference from the \emph{Mix*} corpora is one of surface variety rather than of task. A \emph{MixSNIPS} utterance joins its two parent queries with one of a small fixed set of connectives, as in Figure~\ref{fig:mix-snips-example}, so the seam between the two intents is lexically predictable. \emph{BlendX} varies the connective and allows the second clause to refer back to the first pronominally, which is precisely what its conjunction- and pronoun-based complexity metrics are meant to register.
Experiments show that state-of-the-art multi-intent detection models struggle more on \emph{BlendX} than on \emph{MixATIS} and \emph{MixSNIPS}, suggesting that existing multi-intent benchmarks may underestimate task difficulty.

\paragraph{\textbf{TOP}}~ \citet{gupta-etal-2018-semantic} argued that the standard manner of representing utterances as intent and slot annotations limits the type of models that can be used in the NLU module of typical task-oriented systems.
Instead, \citet{gupta-etal-2018-semantic} proposed a hierarchical representation strategy where slots and intents may be nested, and argue that queries such as
\begin{center}
    \emph{give me driving directions to the eagles game}
\end{center}
ought to be represented as a composition of two intents: \texttt{get\_directions} and \texttt{get\_location}.
As such, \citet{gupta-etal-2018-semantic} introduced the \emph{TOP} (Task Oriented Parsing) dataset, a corpus of 44,783 utterances across 25 intents and 36 slots labeled using a hierarchical annotation scheme.

\par\vspace{\intextsep}\noindent\begin{minipage}{\columnwidth}\centering
    \includegraphics[width=\columnwidth]{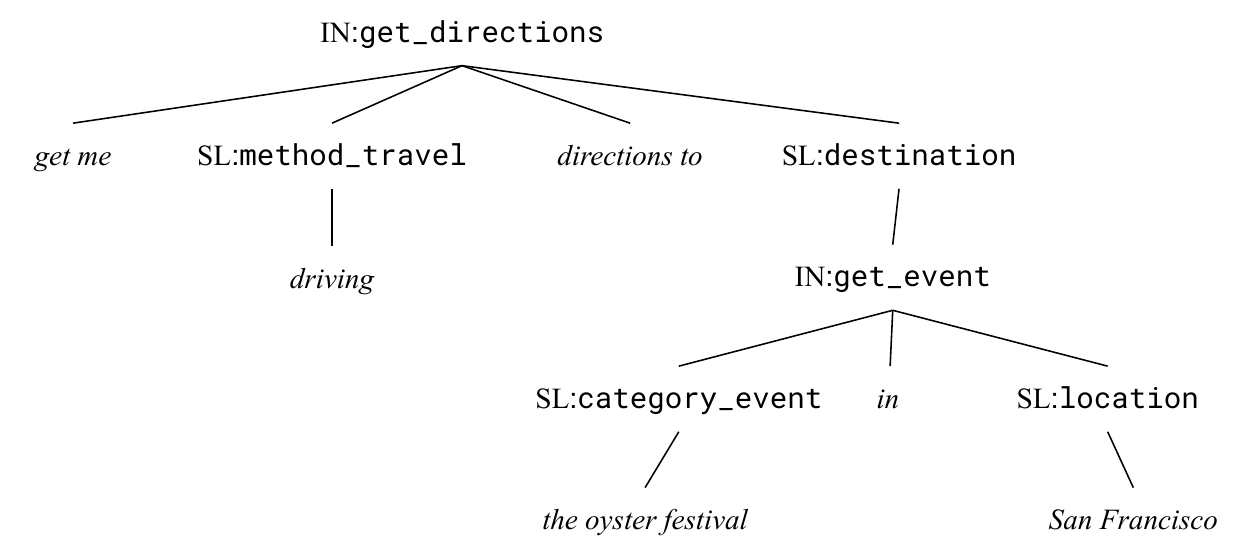}
    \captionof{figure}{Example representation of an utterance in the \emph{TOP} corpus. (IN: and SL: refer to intent and slot, respectively.)}
    \label{fig:top_example_parse}
\end{minipage}\par\vspace{\intextsep}

Figure~\ref{fig:top_example_parse} displays an example annotation for the query \emph{get me driving directions to the oyster festival in San Francisco}. 
This utterance is annotated with two intents.
All utterances in \emph{TOP} have a top-level intent, and we observed that roughly 35\% of all utterances have multiple intents.
As such, \emph{TOP} can be considered a multi-intent dataset; however, all of the multi-intent utterances that we observed have the structure where the nested intent is a direct and only child of a slot annotation, like in the example in Figure~\ref{fig:top_example_parse} where the \texttt{get\_event} intent is a direct, only child of the \texttt{destination} slot.
(In other words, at least some of the nested intents are redundant with respect to their parent slot annotations.)

\paragraph{\textbf{TOPv2}}~ \citet{chen-etal-2020-low-topv2} built off of the \emph{TOP} corpus \citep{gupta-etal-2018-semantic} by adding 72 new intents to create \emph{TOPv2}.
In total, the \emph{TOPv2} corpus consists of roughly 180,000 utterances across 80 intents and has 82 slot types.
Like \emph{TOP}, the \emph{TOPv2} corpus was created by crowdsourcing utterances.
The annotation structure of \emph{TOPv2} uses the same hierarchical style as \emph{TOP}, and we estimate that roughly 16\% of \emph{TOPv2} are multi-intent utterances.
A spoken version of \emph{TOPv2} exists, called \emph{\textbf{STOP}} \citep{tomasello-stop}, which consists of 236,477 speech recordings from 885 unique crowd workers from Amazon Mechanical Turk.


\paragraph{\textbf{HWU-64}}~ The \emph{HWU-64} joint corpus \citep{XLiu.etal:IWSDS2019-hwu} covers a wide variety of intent types, including home automation, travel, and other general categories (e.g., weather queries).
In total, this dataset has 64 intent categories and 54 slot types across 25,716 crowdsourced utterances.
Like \emph{STOP} with \emph{TOPv2}, the \emph{HWU-64} corpus has a spoken language extension, \textbf{\emph{SLURP}} \citep{bastianelli-etal-2020-slurp}, which has 72,277 speech recordings of utterances from \emph{HWU-64} from 177 unique speakers.

The breadth of the \emph{HWU-64} intent inventory comes at a cost to annotation consistency: \emph{HWU-64} exhibits one of the highest estimated label-noise rates among the English-language datasets we audit, with a confident-learning analysis flagging roughly 14\% of its training utterances as candidate label errors.
This is largely because several of its fine-grained intents are difficult even in principle to tell apart.
The most extreme case is the pair \texttt{cooking\_query} and \texttt{cooking\_recipe}: in cross-validation a linear probe confuses the two so thoroughly that essentially every \texttt{cooking\_query} utterance is assigned to \texttt{cooking\_recipe}---for example, \emph{``I need a recipe for chicken thighs for my pressure cooker''} is labeled \texttt{cooking\_query} even though it explicitly asks for a recipe.
A comparable ambiguity separates \texttt{general\_affirm} and \texttt{general\_praise}, where we find identical utterances split across the two labels: \emph{``it's wonderful.''} and \emph{``that's excellent.''} each appear under both intents.
Other frequently conflated pairs include \texttt{general\_greet} versus \texttt{general\_quirky} and \texttt{audio\_volume\_other} versus \texttt{audio\_volume\_up}.
These cases are less outright annotation mistakes than symptoms of an intent scheme whose categories overlap semantically, and they impose a practical ceiling on achievable accuracy---consistent with \emph{HWU-64} being among the least linearly separable of the datasets we examine (a five-fold linear-probe accuracy of 85\%, below the 94\% we measure for \emph{Clinc-150}).

\paragraph{\textbf{CAIS}}~ The \emph{CAIS} (\emph{Chinese Artificial Intelligence Speakers}) corpus is a Chinese language benchmark consisting of 11 intents and 24 slots \citep{cais-cmnet}.
While not explicitly stated in \citet{cais-cmnet}, \emph{CAIS} appears to have been constructed from real, spoken user utterances to a production dialog system. \emph{CAIS} consists of 10,001 samples.
Its intent inventory is small and its domains---music playback, device control, weather, and similar smart-speaker tasks---are narrow relative to the other multilingual corpora in this section, which makes it more comparable in scope to \emph{Facebook} than to broad benchmarks such as \emph{MASSIVE} or \emph{MTOP}.
\emph{CAIS} is also one of the few Chinese-language joint benchmarks in this survey, alongside \emph{FewJoint} and \emph{SMP-ECDT Task 1}, and unlike them it is drawn from deployed-system traffic rather than from crowdsourcing or few-shot construction. Because Chinese is written without word delimiters, the lexical measures we report for \emph{CAIS} are computed over characters and are therefore not directly comparable to the word-level figures for the English corpora.

\paragraph{\textbf{Facebook}}~ Like \emph{Multilingual ATIS} and \emph{Almawave-SLU}, the \emph{Multilingual Task Oriented Dialog} dataset (or \emph{Facebook}, for short) \citep{schuster-etal-2019-cross-lingual} was created to evaluate multilingual transfer learning; it takes its short name from its authors, who were researchers at Facebook.
This dataset consists of 12 intents across three domain areas related to setting alarms, reminders, and querying the weather.
There are 11 slot types.
English language utterances were first gathered by prompting crowd workers to provide example commands or questions they would say to a device capable of handling the three intent categories, then separate crowd workers annotated intent and slot labels for each utterance.
A subset of the English utterances were then translated into Spanish and Thai by native speakers.

\begin{table}[!htbp]
    \centering
    \caption{Non-English joint datasets.}
    \adjustbox{max width=\columnwidth}{
    \begin{tabular}{lccc}
    \toprule
        \textbf{Dataset} & \textbf{Intents} & \textbf{Slots} & \textbf{Language}  \\ 
        \midrule
        \emph{Multilingual ATIS} \citep{multilingual-atis} & 28 & 82 & hi,tr \\ 
        \emph{Facebook} \citep{schuster-etal-2019-cross-lingual} & 12 & 11 & en, es, th \\
        \emph{Almawave-SLU} \citep{almawave} & 7 & 39 & it \\
        \emph{CAIS} \citep{cais-cmnet} & 11 & 24 & zh \\
        \emph{MultiATIS++} \citep{multiatispp} & 18 & 84 & \begin{tabular}[t]{@{}c@{}}de,en,es,fr,hi,\\ja,pt,tr,zh\end{tabular} \\
        \emph{Leyzer} \citep{leyzer} & 186 & 86 & en,es,pl  \\
        \emph{FewJoint} \citep{hou2020fewjoint} & 143 & 205 & zh\\
        \emph{MTOP} \citep{li2021mtop} & 117 & 78 & \begin{tabular}[t]{@{}c@{}}de,en,es,\\fr,hi,th\end{tabular} \\
        \emph{PhoATIS} \citep{JointIDSF-phoatis} & 28 & 82 & vi \\
        \emph{xSID} \citep{van-der-goot-etal-2021-masked-xSID} & 16 & 41 & \begin{tabular}[t]{@{}c@{}}ar,da,de,de-st,en,id,it,\\ja,kk,nl,sr,tr,zh\end{tabular} \\
        \emph{Persian ATIS} \citep{akbari-etal-2023-persian-atis} & --- & --- & fa \\
        \emph{MASSIVE} \citep{fitzgerald-etal-2023-massive} & 60 & 55 & \begin{tabular}[t]{@{}c@{}}51 languages,\\29 families\end{tabular} \\
        \bottomrule
    \end{tabular}}
    \label{tab:other-datasets}
\end{table}

\paragraph{\textbf{MTOP}} Inspired by the nested nature of the queries in many of the \emph{TOP} dataset's utterances, the \emph{MTOP} benchmark was created as a large joint dataset consisting of nested queries in 6 languages: English, Spanish, French, German, Hindi, and Thai \citep{li2021mtop}.
\emph{MTOP}'s utterances cover 11 different domains and 117 intents (each domain ranges between 3 and 27 intents).
There are 78 slot types.
The dataset was constructed in several phases: First, crowd workers were prompted to provide English utterances to a hypothetical system given a certain domain.
Then, to build the non-English versions of \emph{MTOP}, professional translators were used to translate the English utterances into each target language, taking care to maintain all slot information.

\paragraph{\textbf{MASSIVE}}~ The \emph{MASSIVE} (\emph{Multilingual Amazon SLU resource package (SLURP) for Slot-filling, Intent classification, and Virtual assistant Evaluation}) dataset \citep{fitzgerald-etal-2023-massive} is one of the largest and most typologically diverse multilingual NLU datasets available.
\emph{MASSIVE} consists of approximately 1 million parallel labeled utterances spanning 51 languages from 29 language families, 18 domains, 60 intents, and 55 slots.
The dataset was created by professional translators who localized the English-only \emph{SLURP} corpus (the spoken extension of \emph{HWU-64}, discussed above) into 50 additional languages, making \emph{MASSIVE} a fully parallel multilingual corpus.

\paragraph{\textbf{Leyzer}} Another multilingual corpus, the \emph{Leyzer} dataset \citep{leyzer} covers English, Spanish, and Polish. \emph{Leyzer} pushes the limits in terms of imbalanced datasets, and has between one and 672 samples per intent class.
It is also the largest dataset presented in this survey in terms of intents, weighing in at 186 intents covering standard tasks that a hypothetical intelligent device could handle (e.g., news, weather, calendar, web search, etc.).
Each intent belongs to one of 20 domain categories.
\emph{Leyzer} contains 86 slot types, with each domain containing between one and seven slot types. 

The \emph{Leyzer} dataset deviates from most of the datasets discussed in this survey in the manner that its sample utterances are generated.
While most dialog datasets (i.e., task-oriented dialog in general, and intent classification and slot-filling in particular) are generated by humans using crowdsourcing, experts, or user queries, \emph{Leyzer} is generated using grammars.
In particular, experts defined 20 grammars that produced template utterances, which were then filled in with values sampled from lists. (Figure~\ref{fig:grammar-example} shows an example of a basic grammar used by \emph{Leyzer}.)
Grammars were built independently for the non-English languages as well, and thus \emph{Leyzer} is not a parallel corpus (unlike most of the other multilingual corpora discussed in this survey), although there is a small subset of each language's dataset that is parallel among the three languages.

Two properties of \emph{Leyzer} stand out in our measurements.
First, its class imbalance is extreme even by the standards of this survey: the normalized intent entropy of each language partition is roughly 0.83 to 0.85, well below the near-uniform values (close to 1.0) of balanced benchmarks such as \emph{Clinc-150} and \emph{Snips}.
Second, although \emph{Leyzer} is generated by grammars---and might therefore be expected to be noise-free by construction---a confident-learning audit flags between 8\% and 9\% of its utterances as candidate label errors in each of the three languages.
The reason is that many \emph{Leyzer} intents are distinguished not by what the user wants but by which optional slots happen to be realized, so grammar outputs that omit the distinguishing slot become near-identical---or outright identical---across intents.
For example, the identical Polish query \emph{``what will the weather be''} is assigned to both \texttt{Weather\_OpenWeather} and \texttt{Weather\_WeatherTomorrow}; in Spanish, \emph{``show me my Google Drive files''} is labeled \texttt{Gdrive\_OpenGdrive} while the near-identical \emph{``show me my starred Google Drive files''} is labeled \texttt{Gdrive\_ShowFilesWithStar}.
Such collisions arise wherever \emph{Leyzer}'s compositional intent scheme---with intents such as \texttt{Contacts\_EditContactWithNumber} versus \texttt{Contacts\_ShowContactWithNameAndWithNumberAndWithType}---draws intent boundaries that a single utterance need not lexicalize.
In effect, distinctions that are really about slot content are promoted to separate intents, which both inflates the intent inventory and makes the resulting classes hard to tell apart (a five-fold linear-probe accuracy of only 77\% to 79\% across the three languages).


\paragraph{\textbf{FewJoint}}~ The \emph{FewJoint} benchmark is a Chinese language benchmark aimed at evaluating few-shot learners.
The corpus consists of 143 intent categories and 205 slot types over what \citet{hou2020fewjoint} consider to be 59 domains, with 6,694 total utterances.
The utterances were sourced from a mix of real user and crowdsourced utterances.
Examples from \emph{FewJoint} are listed in Table~\ref{tab:fewjoint-examples}.
The domain count is what makes \emph{FewJoint} unusual: at 59 domains over 6,694 utterances it averages barely a hundred utterances per domain, which is by design---the benchmark is built so that a model must be evaluated on domains it has not been trained on, with each domain supplying only a handful of labeled examples.
This inverts the usual arrangement in this survey, where a corpus offers many examples across a few domains, and it makes \emph{FewJoint} closer in spirit to \emph{NLU++} and the \emph{HINT3} corpora, which likewise target the low-resource end of the problem, than to large benchmarks such as \emph{MASSIVE}. Its slot inventory is also the largest of any dataset in this survey, at 205 types.

\begin{CJK*}{UTF8}{gbsn}
\begin{table}[!htbp]
    \centering
    \caption{Example queries from FewJoint \citep{hou2020fewjoint}.}
    \adjustbox{max width=\columnwidth}{\begin{tabular}{ccc}
        \toprule
        \textbf{Domain} & \textbf{Intent} & \textbf{Utterance}\\
        \midrule
        \texttt{message} & \texttt{sendcontacts} & 把王世怀的号码发给乔丽君 \\
        \texttt{email} & \texttt{send} & 发一封邮件给张三，内容是晚上来我家吃饭\\
        \texttt{weather} & \texttt{query} & 合肥今天中午12点温度是多少度\\
        \texttt{cinemas} & \texttt{score\_query} & 大话西游之月光宝盒的评分有多高\\
        \texttt{translation} & \texttt{translation} & 你今天去了哪里用英文怎么讲\\
        \texttt{home} & \texttt{turn\_on\_light} & 打开客厅的吊灯\\
        \bottomrule
    \end{tabular}}
    \label{tab:fewjoint-examples}
\end{table}
\end{CJK*}

\paragraph{\textbf{CSTOP}}~ The \emph{CSTOP} corpus \citep{cstop} is a mixed-language Spanish and English language dataset consisting of \emph{code-switched} (``Spanglish'') queries.
These queries target two domains---weather and device---and belong to one of 19 intents and may contain any of 10 slots.
While there are several other multilingual intent classification and slot-filling corpora (e.g., \emph{Multilingual ATIS}, \emph{Almawave-SLU}, \emph{Multilingual Task Oriented Dialog}, \emph{MultiATIS++}, and \emph{Leyzer}), most of these are derived from existing corpora (i.e., \emph{ATIS} and \emph{Snips}).
The highly similar semantic parsing field also has a code-switching test set \citep{duong-etal-2017-multilingual} for the \emph{NLMaps} corpus \citep{haas-riezler-2016-corpus}, yet their code-switched corpus was constructed from combining two monolingual data sources.
In contrast, \emph{CSTOP} was constructed from the ground up by workers proficient in code-switched Spanish and English.
Examples from \emph{CSTOP} are shown in Table~\ref{tab:cstop-examples}.
\emph{CSTOP} follows the hierarchical annotation style of the other \emph{TOP}-style datasets.
While \emph{CSTOP} contains some utterances with multiple intent annotations, we estimate that only around 6\% have multi-intent annotations, and these appear almost exclusively in \emph{CSTOP}'s weather domain.

\begin{table}[!htbp]
    \centering\caption{Examples from the code-switched language corpus, \emph{CSTOP} \citep{cstop}.}
    \adjustbox{max width=\columnwidth}{\begin{tabular}{ll}
    \toprule
    \textbf{Intent} & \textbf{Utterance}\\
    \midrule
    \texttt{turn\_on} &\emph{activate mi modo de security}\\
    \texttt{open\_homescreen} & \emph{go to la p\'{a}gina de inicio}\\
    \texttt{sleep\_mode} & \emph{pon el dispositivo en sleep mode}\\
    \texttt{open\_resource} & \emph{take me a superframe por favor}\\
    \texttt{mute\_volume} & \emph{pon el speaker en mute}\\
    \texttt{turn\_on} & \emph{quiero grabar usando la smart camera}\\
    \texttt{turn\_off} & \emph{ponle sonido al speaker}\\
    \texttt{set\_brightness} & \emph{increase el brillo to 70\%}\\
    \texttt{maximize\_volume} & \emph{turn up the volumen a su maxima potencia}\\
    \texttt{get\_weather} & \emph{quiero saber el weather despu\'{e}s de las cinco}\\
    \texttt{get\_weather} & \emph{dima cu\'{a}l es el uv index para hoy}\\
    \texttt{get\_weather} & \emph{necesito ponerme un rain jacket}\\
    \texttt{get\_weather} & \emph{tell me the weather para esta tarde}\\
    \bottomrule
    \end{tabular}}
    \label{tab:cstop-examples}
\end{table}

\paragraph{\textbf{xSID}}~ The \emph{xSID} corpus seeks to serve as a benchmark for cross-lingual transfer, providing copious amounts of training data in English, and limited amounts of evaluation data in 13 different languages: Arabic, Chinese, Danish, Dutch, English, German, Indonesian, Italian, Japanese, Kazakh, Serbian, Turkish, and South Tyrolean (each language has 800 evaluation utterances) \citep{van-der-goot-etal-2021-masked-xSID}.
To construct the dataset, the creators of \emph{xSID} sampled data from \emph{Snips} and \emph{Facebook} (both discussed above); these utterances were then translated by experts from English to each target language.
\emph{xSID} consists of 16 intents and 41 slot types.

\begin{table}[!htbp]
    \centering
    \caption{Example utterances from \emph{NLU++} \citep{Casanueva2022-nlu}.}
    \adjustbox{max width=\columnwidth}{\begin{tabular}{lll}
    \toprule
        \textbf{Domain} & \textbf{Intent(s)} & \textbf{Utterance}  \\
    \midrule
         \texttt{hotels} & \texttt{change}, \texttt{booking}, \texttt{room} & \emph{i want to change my room reservation}\\
       \texttt{hotels} & \texttt{request\_info}, \texttt{check\_in} & \emph{can i check in in the morning}\\
         \texttt{hotels} & \texttt{change}, \texttt{booking} & \emph{i need to amend my booking}\\
         \texttt{banking} & \texttt{card}, \texttt{lost\_stolen}, \texttt{credit} & \emph{i lost my mastercard. it was credit i think}\\
         \texttt{banking} & \texttt{when}, \texttt{appointment} & \emph{what time is my appointment tomorrow?}\\
    \bottomrule
    \end{tabular}}
    \label{tab:nlu-examples}
\end{table}

\paragraph{\textbf{NLU++}}~ The \textbf{\emph{NLU++}} \citep{Casanueva2022-nlu} benchmark consists of two individual joint-task datasets: \emph{Banking} (consisting of 48 intents and 13 slots) and \emph{Hotels} (consisting of 40 intents and 14 slots).
Together, these two datasets form \emph{NLU++}, which has 62 intents, 17 unique slots, and a total of 3,080 user-generated utterances, many of which are multi-intent.
The annotation scheme used by \emph{NLU++} is unique: instead of annotating spans to indicate separate intent segments, a set of intent labels is applied to each utterance.
The intent labels can have varying degrees of granularity: for instance, the \texttt{cancel} intent category can be applied to a wide variety of utterances across both the \emph{Banking} and \emph{Hotel} domains, while the \texttt{account} intent applies only to the \emph{Banking} domain. Together, both \texttt{cancel} and \texttt{account} can be applied to utterances like \emph{please cancel my bank account} without the need for an explicit \texttt{cancel\_bank\_account} intent category.

\section{Intent Classification Datasets}\label{sec:intent-classification-datasets}

Intent classification datasets are typically similar to joint intent classification and slot-filling datasets, except they lack slot-filling annotations.
Such datasets are usually designed with the goal of evaluating only classification models, but they can also be used for evaluating tasks like clustering as well.
The datasets discussed in this section are summarized in Table~\ref{tab:intent-classification-datasets}.

\paragraph{\textbf{SMP-ECDT Task 1}}~ The \emph{SMP-ECDT Task 1} dataset consists of 3,736 utterances in Chinese \citep{zhang2019evaluation-smp2017}.
\emph{SMP-ECDT Task 1} has relatively few utterances per intent.
The dataset has 31 intent categories, covering a wide breadth of topics.
Example utterances from this corpus are listed in Table~\ref{tab:smp-ecdt-samples}.
Utterances in \emph{SMP-ECDT Task 1} were sourced from real user utterances.

\begin{CJK*}{UTF8}{gbsn}
\begin{table}[!htbp]
    \centering
    \caption{Example utterances from \emph{SMP-ECDT Task 1} \citep{zhang2019evaluation-smp2017}.}
    \adjustbox{max width=\columnwidth}{\begin{tabular}{cc}
        \toprule
        \textbf{Intent} & \textbf{Utterance}\\
        \midrule
        \texttt{map} & 从二手车市场到八一广场怎么走\\
        \texttt{news} & 最新的新闻播报一下\\
        \texttt{calc} & 123和34的和等于多少\\
        \texttt{telephone} & 打电话给郑艺红\\
        \texttt{flight} & 查一下从南京到上海的航班\\
        \texttt{translation} & 你去哪用英文怎么说\\
        \texttt{stock} & 看一下中式传媒股票的涨跌幅\\
        \texttt{weather} & 帮我查一下赣州的天气明天呢\\
        \bottomrule
    \end{tabular}}
    \label{tab:smp-ecdt-samples}
\end{table}
\end{CJK*}

\paragraph{\textbf{Outlier Collection}}~ The \emph{Outlier Collection} \citep{larson-etal-2019-outlier} was not designed to be a benchmark for comparing intent classification models, but rather a way to compare data collection methods. The \emph{Outlier Collection} consists of three datasets, \emph{Same}, \emph{Random}, and \emph{Unique}, but each dataset contains the same 10 intent classes. The difference between the three sets is the manner in which they were crowdsourced.

All three were collected over three rounds of \emph{paraphrase} crowdsourcing prompts on Amazon Mechanical Turk, and differ only in how the seed phrases for each round were chosen: \emph{Same} holds them constant, \emph{Random} samples them at random from the previous round, and \emph{Unique} uses outlier detection to pick the most distinctive (but still semantically correct) utterances from the previous round.
In this way, the manner in which the \emph{Unique} version of \emph{Outlier} is generated is inspired by \citet{negri-etal-2012-chinese-whispers}, which uses a method similar to the ``Chinese whispers'' or ``telephone'' game of iterative paraphrase generation.
All 10 intents for each \emph{Outlier} dataset belong to the banking domain, and there is a large number of samples per intent, with a total of 6,079 samples across all intents (in the case of the \emph{Unique} dataset).

\paragraph{\textbf{Clinc-150}}~ The \emph{Clinc-150} dataset \citep{larson-etal-2019-evaluation} was designed to target several dimensions of intent classification evaluation. First, to test the limits of intent classifiers by providing a large number (150) of intent classes. These intent categories each belong to one of 10 domains, including \emph{banking}, \emph{travel}, \emph{kitchen \& dining}, and \emph{work}. The large number of intents in \emph{Clinc-150} stands in contrast with older datasets, like \emph{Snips} (7 intents) and \emph{ATIS} (22 intents). 

Data for \emph{Clinc-150} was collected by crowdsourcing on Amazon Mechanical Turk, using \emph{paraphrase} and \emph{scenario} prompts that asked workers either to paraphrase an existing utterance from an intent or to respond to a scenario that would lead them to express it.
Table~\ref{tab:clinc150} lists example utterances from the \emph{Clinc-150} dataset.

\begin{table*}[tbp]
    \centering
    \caption{Sample utterances from the \emph{Clinc-150} dataset \citep{larson-etal-2019-evaluation}.}
    \scalebox{0.77}{ 
    \begin{tabular}{ccl}
    \toprule
        \textbf{Domain} & \textbf{Intent} & \textbf{Utterance} \\
        \midrule
        \texttt{credit\_cards} & \texttt{card\_declined} & \emph{my card was rejected at shakey's and i am wondering why}\\
        \texttt{kitchen\_and\_dining} & \texttt{restaurant\_reservation} & \emph{i need a table for 3 at 5pm at andrea's steakhouse under wheeler}\\
        \texttt{auto\_and\_commute} & \texttt{traffic} & \emph{what is the traffic like on the way to north shore}\\
        \texttt{utility} & \texttt{time} & \emph{what time is in over there in pacific standard time}\\
        \texttt{travel} & \texttt{flight\_status} & \emph{could you tell me the status of flight dl123}\\
        \texttt{work} & \texttt{schedule\_meeting} & \emph{i'd like to schedule a meeting room from 1:00 pm until 2:00 pm}\\
        \texttt{meta} & \texttt{change\_language} & \emph{please respond to me in english from now on}\\
        \texttt{home} & \texttt{shopping\_list\_update} & \emph{i'm out of kleenex so will you put that on my shopping list}\\
        \texttt{small\_talk} & \texttt{are\_you\_a\_bot} & \emph{tell me if you are a human or are a computer}\\
        \texttt{banking} & \texttt{transfer} & \emph{move 100 dollars from my savings to my checking}\\
        \midrule
        --- & \emph{out-of-scope} & \emph{how long does it take to become an architect}\\
        --- & \emph{out-of-scope} & \emph{is the united states a democracy}\\
        --- & \emph{out-of-scope} & \emph{how much over will overdraft protection cover}\\
        --- & \emph{out-of-scope} & \emph{what were the top stories this week}\\
        --- & \emph{out-of-scope} & \emph{how much has microsoft's stock changed over the last year}\\
        
    \bottomrule
    \end{tabular}}
    \label{tab:clinc150}
\end{table*}

The second motivation---and main novelty---of the design of \emph{Clinc-150} was to include numerous \emph{out-of-scope} utterances, which are intended to test an intent classifier's ability to distinguish queries that belong to the 150-intent \emph{in-scope} set, and those that do not. The same notion appears in the literature as \emph{out-of-distribution} \citep{hendrycks-ood}, \emph{out-of-domain} \citep{tur2014detecting, tan-etal-2019-domain, zheng-ood-2020}, and \emph{out-of-application} \citep[e.g.][]{bohus-rudnicky-2005-sorry} testing. Distinctively, the \emph{Clinc-150} out-of-scope set draws utterances from within the 10 in-scope domains as well as from outside them.

Beyond its scale and out-of-scope design, \emph{Clinc-150} is characterized by balanced and comparatively clean annotations.
It exhibits one of the lower estimated label-noise rates among the datasets we audit: a confident-learning analysis flags only 0.6\% of its training utterances, versus 2.3\% for the similarly fine-grained \emph{Banking-77}.
The residual confusions, moreover, correspond to genuinely fine semantic distinctions rather than apparent annotation errors---for instance, a linear probe most often confuses \texttt{pto\_used} with \texttt{pto\_balance}, and \texttt{todo\_list} with \texttt{todo\_list\_update}.
\emph{Clinc-150} also remains a standard benchmark for LLM-based intent detection and discovery, including recent training-free few-shot approaches \citep{rodriguez-2024-intentgpt}, and its dedicated out-of-scope split continues to make it a reference point for evaluating whether models can abstain on queries beyond their supported intents.


\paragraph{\textbf{Banking-77}}~
Where \emph{Clinc-150} spreads its intents across ten disparate domains, the \emph{Banking-77} dataset \citep{Casanueva2020} confines all 77 of its intents to a single one---banking---on the rationale that telling closely-related intents apart is the harder task.
Its data was collected by crowdsourcing\footnote{Determined via author correspondence.} in a manner similar to \emph{Clinc-150}.

This design does make the task measurably harder: a simple linear probe over multilingual sentence embeddings classifies \emph{Clinc-150} intents at 94.4\% accuracy but reaches only 87.8\% on \emph{Banking-77}, despite the latter having roughly half as many intent classes, and \emph{Banking-77}'s intents are barely separated in embedding space (cosine silhouette of 0.10, versus 0.21 for \emph{Clinc-150}).
The residual confusions fall predictably along semantically adjacent intents; for example, 17\% of \texttt{virtual\_card\_not\_working} utterances are assigned by the probe to \texttt{getting\_virtual\_card}.

The very fine granularity that makes \emph{Banking-77} challenging, however, also renders it prone to annotation inconsistency: \citet{ying-thomas-2022-label} estimate that roughly 14\% of its training utterances may be mislabeled, and report that removing the suspect examples improves downstream classification.
Our own audit corroborates the kind of error rather than its magnitude: confident learning flags a far smaller share of \emph{Banking-77} (2.3\%, Table~\ref{tab:corpus-diagnostics}), since the two methods target different notions of a mislabeled example, but the audit surfaces numerous near-identical utterances that carry different intent labels.
For instance, \emph{``How can I get new card?''} is labeled \texttt{contactless\_not\_working} while the near-duplicate \emph{``how do I get a new card''} is labeled \texttt{order\_physical\_card}; similarly, \emph{``Is there a way to reset my PIN?''} is labeled \texttt{pin\_blocked} whereas \emph{``How can I reset my PIN?''} is labeled \texttt{change\_pin}.
Such cases are difficult to fault the annotators for---the underlying intent categories genuinely overlap---but they place an upper bound on the accuracy any classifier can achieve, a theme we return to in Section~\ref{sec:discussion-future-work}.
Finally, \emph{Banking-77} is released as a train/test partition only, with no designated validation split.
As a result, researchers construct their own validation sets in varying (and often unreported) ways, which undercuts the strict comparability of the accuracies reported against this benchmark.

\begin{table*}[tbp]
    \centering
    \caption{Intent Classification datasets.}
    \scalebox{0.87}{
    \begin{tabular}{lccccc}
    \toprule
        \textbf{Dataset} & \textbf{Intents} & \textbf{\# Utterances} & \textbf{Source} & \textbf{Lang.} & \textbf{License} \\
        \midrule
        \emph{SMP-ECDT Task 1} \citep{zhang2019evaluation-smp2017} & 31 & 3,736 & Users & zh & ---\\
        \emph{Outlier} \citep{larson-etal-2019-outlier} & 10 & 6,079 & Crowd & en & \textsc{cc by-nc 3.0} \\
         \emph{Clinc-150} \citep{larson-etal-2019-evaluation} & 150 & 23,700 & Crowd & en & \textsc{cc by 3.0} \\ 
         \emph{Banking-77} \citep{Casanueva2020}& 77 & 13,082 & Crowd & en & \textsc{cc by-nc 4.0}\\ 
         \emph{ACID} \citep{ACID2020} & 174 & 22,172 & Users & en & ---\\ 
         \emph{Curekart} \citep{arora-etal-2020-hint3} & 28 & 1,590 &  Experts, Users & en & \textsc{ODbl 1.0}\\
         \emph{Powerplay11} \citep{arora-etal-2020-hint3} & 59 & 1,454 & Experts, Users & en & \textsc{ODbl 1.0}\\
         \emph{SOFMattress} \citep{arora-etal-2020-hint3} & 21 & 710 & Experts, Users & en & \textsc{ODbl 1.0}\\
                  \emph{ROSTD} \citep{gangal-aaai2020-rostd} & --- & 4,590 & Experts & en & --- \\
         \emph{Redwood} \citep{Larson2022-redwood} & 451 & 62,216 & Crowd, Mixed & en & --- \\
         \emph{StanfordLU} \citep{hou-etal-2021-stanfordlu} & --- & 8,038 & Crowd & en & --- \\
         \emph{MInDS-14} \citep{gerz-etal-2021-minds14} & 14 & --- & Crowd & 14 varieties & \textsc{cc-by 4.0} \\
         \emph{CREDIT16} \citep{song-etal-2023-aktify} & 16 & 4,384 & Users & en & --- \\
         \bottomrule
    \end{tabular}}
    \label{tab:intent-classification-datasets}
\end{table*}

\paragraph{\textbf{ACID}}~ The \emph{Amfam Chatbot Intent Dataset} (\emph{ACID}) \citep{ACID2020} follows \emph{Banking-77} by providing a large number of intents from a single domain.
\emph{ACID} has 174 intents all belonging to the insurance domain, more than double the number of intents from \emph{Banking-77}.
Unlike \emph{Clinc-150} and \emph{Banking-77}, \emph{ACID} consists of utterances sourced from real customer interactions with human customer service representatives from American Family Insurance (Amfam) company.

\paragraph{\textbf{CREDIT16}}~
The \emph{CREDIT16} dataset \citep{song-etal-2023-aktify} is an industry corpus from Aktify Inc., introduced as a real-world validation benchmark for the PCMID (\emph{Prototypical Contrastive Multi-Intent Detection}) framework, which learns multiple semantic representations per utterance via supervised prototypical contrastive learning.
\citet{song-etal-2023-aktify} describe it as a private commercial corpus in the credit repair domain, with 16 intent labels over 4,384 utterances.
Unlike most intent classification datasets, \emph{CREDIT16} is \emph{multi-intent}, and its utterances are long by the standards of this survey, being conversational SMS turns rather than short directed queries.
Its interaction type also differs from the other company-specific corpora discussed above: whereas \emph{ACID} and the \emph{HINT3} datasets capture inbound customer service, \emph{CREDIT16} is predominantly \emph{outbound} sales and lead qualification, and its label set is organized around advancing or terminating a sales conversation---the most frequent labels include do-not-contact, not-interested, cost-too-high, and chose-competitor.

The construction of the label set is unusually well documented, and is instructive for the discussion of annotation schemes in Section~\ref{sec:discussion-future-work}.
\citet{song-etal-2023-aktify} began from roughly 130 candidate intents, pre-labeled utterances using a binary classifier per intent, and had three contracted linguists annotate them. Any label with fewer than 100 examples was then dropped or folded into a semantically adjacent one---for instance, the various reasons a user gives for asking not to be contacted were collapsed into a single category---and the annotators revised their labels under the resulting 16-category scheme.
All personally identifying information was removed automatically and then checked manually, with names replaced by placeholders.

\paragraph{\textbf{HINT3}}~ The \emph{HINT3} corpora \citep{arora-etal-2020-hint3} consist of 3 datasets each from a single domain. These datasets are named \textbf{\emph{SOFMattress}} (21 intents), \textbf{\emph{Curekart}} (28 intents), and \textbf{\emph{Powerplay11}} (59 intents). Notably, the \emph{HINT3} datasets also include a sizeable number of \emph{out-of-scope} test samples. Unique among the datasets discussed thus far, the \emph{HINT3} datasets have training data sourced from domain experts who imitate real users, based on historical user queries from deployed systems. The \emph{HINT3} test sets---including the \emph{out-of-scope} data---consist of user data from deployed systems.
This construction leaves a clear quantitative signature: because the training and test data come from genuinely different sources rather than from a single annotation effort split in two, the \emph{HINT3} corpora exhibit by far the highest lexical novelty of any dataset we measure, with 28\% (\emph{Curekart}), 32\% (\emph{SOFMattress}), and 43\% (\emph{Powerplay11}) of test tokens unseen during training---an order of magnitude above the crowdsourced benchmarks discussed above, where the corresponding figure is often below 1\%.
For evaluating whether a model generalizes to real user language, this property arguably matters more than corpus size.

\paragraph{\textbf{ROSTD}}~ \citet{gangal-aaai2020-rostd} created \emph{ROSTD} (\emph{Real Out-of-Domain Sentences from Task-Oriented Dialog}) as a companion out-of-domain benchmark to the \emph{Facebook} corpus \citep{schuster-etal-2019-cross-lingual}.
In this way, \emph{ROSTD} is similar to the out-of-scope set provided with the \emph{Clinc-150} benchmark.
To create \emph{ROSTD}, several workers first familiarized themselves with the in-domain intent categories from \emph{Facebook}, then worked to produce realistic utterances that are out-of-domain with respect to \emph{Facebook}'s intent categories.
\emph{ROSTD} consists of 4,590 utterances.

\paragraph{\textbf{Redwood}}~ \citet{Larson2022-redwood} observed that many of the published intent classification and joint-task datasets have overlapping intents (e.g., \emph{Snips} and \emph{Clinc-150} both have intents for asking about the weather), as well as intent categories that are unique to a particular dataset.
\citet{Larson2022-redwood} then developed automated tools to detect ``colliding'' intents between datasets (i.e., intent categories from individual datasets that overlap semantically with intents from other datasets) in order to join 12 intent classification datasets together into a single intent classification dataset, \emph{Redwood}. The \emph{Redwood} dataset consists of 451 intents, with each intent consisting of at least 50 samples. The 12 individual datasets that were used to build \emph{Redwood} are: \emph{ACID}, \emph{Clinc-150}, \emph{MTOP}, \emph{Banking-77}, \emph{HWU-64}, \emph{MetalWOz}, \emph{DSTC-8}, \emph{ATIS}, \emph{Outlier}, \emph{Snips}, \emph{Jobs640}, and \emph{Talk2Car}. \emph{Redwood} also includes intent categories not found in those 12 datasets; the authors crowdsourced this data.
Our vocabulary-overlap measurements confirm that this aggregation succeeds in subsuming its sources: 94\% of \emph{Banking-77}'s training vocabulary, 88\% of \emph{Clinc-150}'s, and 94\% of \emph{MInDS-14}'s recur in \emph{Redwood}, while \emph{Redwood} retains a substantially larger vocabulary than any of them (11,595 word types).
A practical consequence is that \emph{Redwood} and its constituent datasets should not be treated as independent benchmarks.

Merging many source corpora into a single label space, however, introduces a characteristic form of annotation conflict.
Because short, generic utterances recur across many of the constituent datasets, \emph{Redwood} contains numerous instances of identical utterances assigned to different intents---an artifact that the automated ``collision'' detection, designed to \emph{merge} semantically overlapping intents, does not fully eliminate.
For example, in the copy we analyze the utterance \emph{``can you help me''} appears under four distinct \emph{Redwood} intents, and \emph{``i need some help''} under two.
Our confident-learning audit flags roughly 6\% of \emph{Redwood}'s training utterances as candidate label errors, and its most-confused intent pairs are conflated at very high rates (for one pair, a linear probe assigns 85\% of one intent's utterances to another), reflecting both this residual duplication and the inherent difficulty of discriminating 451 fine-grained intents.
Dataset-merging is thus an efficient way to scale up the intent label space, but it tends to inherit and compound the ambiguities of its sources.

\paragraph{\textbf{StanfordLU}}~
The \emph{StanfordLU} dataset \citep{hou-etal-2021-stanfordlu} is a re-annotation of the \emph{Stanford Multi-Domain} dialog corpus \citep{eric-etal-2017-key-value} for \emph{multi-label} intent detection, where a single utterance may be assigned more than one intent label.
The source corpus is a set of 3,031 in-car assistant dialogs covering calendar scheduling, point-of-interest navigation, and weather retrieval; it is a turn-based dialog dataset rather than an intent classification benchmark, and so is not cataloged in this survey in its own right (see Section~\ref{sec:other-datasets}).
The dataset covers 3 domains---\emph{Schedule}, \emph{Navigate}, and \emph{Weather}---and consists of 8,038 utterances.
This multi-label annotation scheme distinguishes \emph{StanfordLU} from most datasets in this survey, which assume each utterance carries exactly one intent.
\citet{hou-etal-2021-stanfordlu} introduce \emph{StanfordLU} alongside a few-shot approach that learns universal intent-thresholding strategies on data-rich source domains and adapts them via nonparametric calibration to low-resource target domains.

\paragraph{\textbf{MInDS-14}}~
The \emph{MInDS-14} (\emph{Multilingual Intent Detection from Spoken Data}) corpus \citep{gerz-etal-2021-minds14} is a multilingual benchmark for intent detection from \emph{spoken} audio.
The corpus covers 14 language varieties---Czech, German, Australian English, American English, British English, Spanish, French, Italian, Korean, Dutch, Polish, Portuguese, Russian, and Chinese---and consists of 14 intent categories drawn from the e-banking domain.
Each instance includes an audio recording, a transcription, an English translation, and an intent label; there are no slot annotations.
\emph{MInDS-14} was designed to support evaluation of multilingual and cross-lingual intent detection from speech, and \citet{gerz-etal-2021-minds14} find that combining machine translation with multilingual sentence encoders (e.g., LaBSE) yields strong cross-lingual intent detectors across target languages.

\section{Slot-Filling Datasets}\label{sec:slot-filling-datasets}
This section surveys slot-filling datasets, which are summarized in Table~\ref{tab:slot_filling_datasets}.
These are typically datasets that consist of utterances with only slot-level annotations.


\paragraph{\textbf{MIT Collection}}~ The \emph{MIT Collection} \citep{mit-asgard-collection} consists of two datasets, \emph{\textbf{MIT Movie}} and \emph{\textbf{MIT Restaurant}}.
The former dataset consists of utterances directed at a hypothetical movie database, while the latter is composed of phrases targeting a restaurant search application. 
Two crowdsourcing strategies were used to generate utterances for both datasets. In the first, called \emph{frame-based}, crowd workers write utterances around one or more slot values they are given. The second, called \emph{free-form}, asks workers to generate queries and does not provide any pre-defined slot values (workers are free to make up their own slot values, or none). 
The \emph{MIT Movie} dataset contains 12,218 samples and 12 slots, and the \emph{MIT Restaurant} dataset has 9,181 samples and 8 slots. 

\paragraph{\textbf{Jaech Collection}}~ The \emph{Jaech Collection} \citep{Jaech+2016} consists of four datasets, all of which consist of queries targeting travel and restaurant apps. These datasets are named \textbf{\emph{Airbnb}}, \textbf{\emph{United}} (i.e., United Airlines), \textbf{\emph{Greyhound}}, and \textbf{\emph{OpenTable}}, and the corresponding app actions that a hypothetical user could take are booking lodging, booking flights, booking bus travel, and reserving a table at a restaurant. The creators of the \emph{Jaech Collection} were motivated to develop these four datasets in order to evaluate multi-task learning slot-filling models. 

Data for the \emph{Jaech Collection} was collected using crowdsourcing. With the exception of the \emph{United} dataset, users were asked to pretend they were interacting with a friend (as opposed to an artificially intelligent app) in an attempt to prompt the crowd workers to use more natural language.
The dataset creators also made an effort to inject a diverse array of slot values into the dataset by sampling values from curated entity lists.
In contrast to most of the datasets discussed thus far, the \emph{Jaech Collection} contains sample utterances from \emph{non-root} states of a dialog.
Contextual information is not provided in the dataset, however, and thus models trained on the \emph{Jaech Collection} must rely solely on the textual features of single utterances.

\begin{table*}[tbp]
    \centering
    \caption{Summary of slot-filling datasets. All datasets listed in this table consist of English utterances.}
    \scalebox{0.95}{ 
    \begin{tabular}{lcccc}
    \toprule
        \textbf{Dataset} & \textbf{Slots} & \textbf{\# Utterances} & \textbf{Source} & \textbf{License} \\
        \midrule
        \emph{MIT Movie} \citep{mit-asgard-collection} & 12 & 12,218 & Crowd & --- \\ 
        \emph{MIT Restaurant} \citep{mit-asgard-collection} & 8 & 9,181 & Crowd & --- \\ 
        \emph{Airbnb} \citep{Jaech+2016} & 11 & 4,663 & Crowd & --- \\
        \emph{Greyhound} \citep{Jaech+2016} & 13 & 4,950 & Crowd & --- \\
        \emph{OpenTable} \citep{Jaech+2016} & 6 & 3,146 & Crowd & --- \\
        \emph{United} \citep{Jaech+2016} & 12 & 20,696 & Crowd & --- \\
        \emph{Restaurants-8k} \citep{coope-etal-2020-span} & 5 & 11,929 & Users & \textsc{cc-by 4.0} \\ 
        \emph{Pizza} \citep{pizzaDataset} & 10 & 2,458,151 & Generated, Crowd & \textsc{cc-by-nc 4.0} \\
        \emph{Burrito} \citep{rubino2022crosstop-foodordering} & 11 & 10,173 & Generated & \textsc{cc-by-nc 4.0} \\
        \emph{Sub} \citep{rubino2022crosstop-foodordering} & 8 & 10,161 & Generated & \textsc{cc-by-nc 4.0} \\
        \emph{Coffee} \citep{rubino2022crosstop-foodordering} & 9 & 104 & Generated & \textsc{cc-by-nc 4.0} \\
        \emph{Burger} \citep{rubino2022crosstop-foodordering} & 9 & 161 & Generated & \textsc{cc-by-nc 4.0} \\
        \bottomrule
    \end{tabular}}
    \label{tab:slot_filling_datasets}
\end{table*}

\paragraph{\textbf{Restaurants-8k}}~ In contrast to the \emph{Jaech Collection}, the \emph{Restaurants-8k} dataset \citep{coope-etal-2020-span} contains contextual annotations indicating which slots were requested by the system in a dialog.
For instance, the utterance \emph{``for just one person''} has ``one person'' annotated with the \texttt{people} slot, but also has the \texttt{people} labeled as a requested slot.
The requested slot annotations thus provide contextual information to help disambiguate slot predictions (e.g., ``7'' could be the number of people or the time for booking a restaurant).
The precise dialog turns are not annotated or included in the dataset, however, nor are system responses provided. 
The \emph{Restaurants-8k} dataset contains 8,198 training and 3,731 test utterances, and was sourced from real users interacting with a deployed dialog system in the restaurant booking domain.

\paragraph{\textbf{Pizza}}~ 
The \emph{Pizza} dataset \citep{pizzaDataset} is a large corpus consisting of 2,456,446 generated training utterances and 1,705 human-generated test and validation utterances.
The synthetic utterances were produced using human-crafted templates.
All samples in the \emph{Pizza} corpus have been annotated in the \emph{TOP} hierarchical fashion, and all utterances in the corpus are related to ordering food and drink from a hypothetical pizza restaurant.

\paragraph{\textbf{FoodOrdering Collection}}~
The \emph{FoodOrdering} collection was introduced in \citet{rubino2022crosstop-foodordering} and consists of 4 distinct datasets: \textbf{\emph{Burrito}}, \textbf{\emph{Sub}}, \textbf{\emph{Coffee}}, and \textbf{\emph{Burger}}, all of which are similar to the \emph{Pizza} dataset (described previously) and consist of \emph{TOP}-style annotated utterances directed at hypothetical food ordering scenarios.
Like \emph{Pizza}, the \emph{Burrito} and \emph{Sub} datasets have synthetically-generated training data, while the evaluation data of all four \emph{FoodOrdering} corpora is crowd-generated.
\emph{Coffee} and \emph{Burger} have no training data, and are intended to be used as evaluation benchmarks for zero-shot learners.

%

\section{Other Datasets}\label{sec:other-datasets}

In this section, we highlight datasets that do not belong to any of the three discussed dataset categories (intent classification, slot-filling, and joint intent classification and slot-filling), but are nonetheless relevant to the task-oriented dialog tasks of intent classification and slot-filling. The datasets discussed in this section are relevant because they contain dialog-style queries, commands, or utterances typically directed toward automated systems.


\subsection{Turn-Based Dialog Datasets}

While this survey focuses on intent classification and slot-filling datasets, turn-based datasets---which include dialogs consisting of a sequence of utterances between two parties---are also relevant.
For instance, the \emph{Redwood} intent classification dataset (discussed in Section~\ref{sec:intent-classification-datasets}) incorporated the ``root'' utterances from several turn-based dialog datasets, including \emph{MetalWOz} \citep{lee2019multi-domain-metalwoz} and \emph{SGD} \citep{rastogi2020towards-sgd-dstc8}. 
The \emph{MetalWOz} dialog corpus consists of 37,884 dialogs, and the root-level utterances are distributed across 51 intents (according to \citet{Larson2022-redwood}).
The \emph{Schema-Guided Dataset} (or \emph{SGD}) consists of 16,142 dialogs collected using crowdsourcing, and \citet{Larson2022-redwood} estimate that the root-level utterances are distributed across 34 intents.
Other turn-based dialog datasets include \emph{MultiWOZ} \citep{budzianowski-etal-2018-multiwoz} (and its extensions \emph{MultiWOZ 2.1} \citep{eric-etal-2020-multiwoz-2.1}, \emph{MultiWOZ 2.2} \citep{zang-etal-2020-multiwoz-2.2}, \emph{MultiWOZ 2.3} \citep{multiwoz-2.3}, and \emph{MultiWOZ 2.4} \citep{Ye2021-oy-multiwoz-2.4}), \emph{M2M} \citep{Shah2018-vn-m2m}, \emph{Frames} \citep{el-asri-etal-2017-frames}, and \emph{WOZ 2.0} \citep{wen-etal-2017-network-woz-2.0}.
Less task-oriented and more open-ended dialog datasets are also valuable, for instance the root-level utterances from the open-ended crowdsourced dialog dataset of \citet{vertanen_aacdialogue} were used as out-of-domain data in \emph{Redwood} \citep{Larson2022-redwood}.

\citet{gretz-etal-2023-vira} introduce \emph{VIRADialogs}, a dataset of over 8,000 real conversations between users and VIRA, a chatbot designed to address questions and concerns surrounding COVID-19 vaccine hesitancy.
Unlike most datasets discussed in this survey, \emph{VIRADialogs} is designed for the task of \emph{intent discovery}---identifying emergent intent categories from unlabeled conversational data---rather than intent classification over a fixed, predefined label set.
This distinction is noteworthy because intent discovery is arguably a prerequisite for building intent classification datasets, and it is a relatively underexplored task in the literature.
The authors accompany the dataset with an automatic evaluation framework that leverages VIRA's existing intent classifier to assess the quality of newly discovered intents, and report that baseline intent discovery approaches struggle on \emph{VIRADialogs}.


\subsection{Short Text Classification Corpora}

Intent classification is inherently a text classification problem, thus corpora used to evaluate text classifiers are relevant here.
One widely-used text classification corpus is the TREC corpus \citep{li-roth-2002-learning-trec}, which consists of natural language queries taken from the 2001 Text Retrieval Conference (TREC) Question Answering challenge.
The TREC corpus has two target label sets, one consisting of six question \emph{types}, each pertaining to the intended answer content sought after in the question.
For example, \emph{location} and \emph{quantity} are both question types.
The second target label set consists of 50 fine-grained question categories (e.g., the \emph{location} question type has the fine-grained sub-categories of \emph{city}, \emph{country}, \emph{mountain}, \emph{state}, and \emph{other}). The TREC corpus consists of 5,952 questions.
A question type is not quite an intent, however, and the distinction matters when TREC is used as a stand-in for an intent classification benchmark. A question type classifies the \emph{answer} the question seeks---whether the expected response is a location, a quantity, a person, and so on---whereas an intent classifies the \emph{action} the user wants a system to take. Two questions with different intents can share a question type, and a single intent can be expressed as a question of several types. TREC labels are therefore a property of the question's semantics rather than of any downstream system behavior, which is why the corpus exercises the classifier without exercising the task.

Other widely used text classification corpora include Movie Reviews \citep{pang-lee-2005-seeing-rotten-tomatoes-movie-reviews}, SST \citep{socher-etal-2013-recursive}, etc., but these are sentiment analysis datasets consisting of product or movie reviews meant to communicate information to other humans rather than utterances directed toward a non-human system.


\subsection{Semantic Parsing and Natural Language to Database Corpora}
The NLP task of semantic parsing aims to convert human language into a machine-readable logical representation (an example application is converting a natural language utterance to a database query).
Semantic parsing has substantial overlap with slot-filling and joint modeling, but we discuss semantic parsing datasets separately here since they are not typically used in dialog evaluation. 

Such datasets include \emph{GeoQuery} \citep{geoquery} and its extensions to Spanish, Japanese, and Turkish \citep{wong-mooney-2006-learning} and to German, Greek, and Thai \citep{jones-etal-2012-semantic}, \emph{Jobs640} \citep{jobs640}, \emph{Restaurants} \citep{tang-mooney-2000-automated}, \emph{Free917} \citep{cai-yates-2013-large}, \emph{WebQuestions} \citep{berant-etal-2013-semantic}, \emph{NLMaps} \citep{haas-riezler-2016-corpus}, \emph{NNLIDB} \citep{brad-etal-2017-dataset}, \emph{Scholar} \citep{iyer-etal-2017-learning}, \emph{MAS} \citep{li-jagadish-academic-sql-2014}, \emph{IMDB} and \emph{Yelp} \citep{yaghmazadeh-et-al-sql-2017}, \emph{WikiSQL} \citep{zhong-wikisql}, \emph{Advising} \citep{finegan-dollak-etal-2018-improving}, \emph{Spider} \citep{yu-etal-2018-spider}, \emph{NL2Bash} \citep{lin-etal-2018-nl2bash}, and \emph{Talk2Car} \citep{deruyttere-etal-2019-talk2car}.

\section{Discussion and Future Directions}\label{sec:discussion-future-work}

In this section we reflect on the nature of the datasets surveyed above. We identify several areas that are ripe for future research directions.

Before turning to those directions, Table~\ref{tab:dataset-guide} offers a quick-reference guide to selecting datasets from this survey according to common evaluation needs.
We stress that the entries within a row are options rather than a recommended suite.

\begin{table}[!htbp]
\centering
\small
\adjustbox{max width=\columnwidth}{\begin{tabular}{@{}>{\raggedright\arraybackslash}p{3.0cm}@{\hspace{4pt}}>{\raggedright\arraybackslash}p{5.0cm}@{}}
\toprule
\textbf{I need\ldots} & \textbf{Recommended datasets} \\
\midrule
Multilingual coverage      & \emph{MASSIVE}, \emph{MultiATIS++}, \emph{MInDS-14} \\
Spoken audio input         & \emph{SLURP}, \emph{MInDS-14} \\
Multi-intent detection     & \emph{BlendX}, \emph{MixATIS}/\emph{MixSNIPS}, \emph{NLU++}, \emph{CREDIT16} \\
Multi-label intents        & \emph{StanfordLU}, \emph{NLU++} \\
Real user utterances       & \emph{ACID}, \emph{CREDIT16} \\
Large scale / many domains & \emph{Clinc-150}, \emph{MASSIVE}, \emph{TOP} \\
Fine-grained intents       & \emph{Banking-77}, \emph{Clinc-150} \\
Challenge / robustness     & \emph{NLU++}, \emph{BlendX} \\
Joint intent + slots       & \emph{ATIS}, \emph{Snips}, \emph{MASSIVE}, \emph{SLURP}, \emph{MultiATIS++} \\
Low-resource languages     & \emph{PhoATIS}, \emph{Persian ATIS}, \emph{MInDS-14} \\
LLM / zero-shot evaluation & Any dataset with a fixed held-out test set; prefer \emph{Clinc-150}, \emph{Banking-77}, \emph{MASSIVE}---but weigh the benchmark-contamination caveat of Section~\ref{subsec:llm-era} \\
\bottomrule
\end{tabular}}
\caption{Dataset selection guide. For each common evaluation need, the most suitable corpora from this survey are listed. Datasets with overlapping strengths appear in multiple rows. Entries within a row are alternatives, not a suite to be used together: some pairs (notably \emph{BlendX} and \emph{MixATIS}/\emph{MixSNIPS}, which derive from the same ancestors) share nearly all of their vocabulary---see the discussion of cross-corpus redundancy in Section~\ref{sec:discussion-future-work}.}
\label{tab:dataset-guide}
\end{table}



\subsection{Gaps in the Current Landscape}

\textbf{Language representation.} Within NLP there is a risk of focusing solely on the English language.
There are by now ample non-English datasets for benchmarking the joint task, but many were produced by automatic translation and so may under-represent realistic language phenomena, and non-English corpora for intent classification or slot-filling alone remain far scarcer.

\textbf{Provenance.} Most of the datasets surveyed here use crowdsourcing or generation to produce utterances, and many of the non-English ones are translations of a parent corpus.
Derived datasets appear elsewhere in NLP as well (e.g., ASR \citep{Chan2021SpeechStewSM}, open information extraction \citep{solawetz-larson-2021-lsoie}), but the intent classification and slot-filling landscape---especially for joint datasets---has become densely populated with them.
Far fewer corpora consist of utterances from real users, and whether models trained on real user data outperform those trained on generated or derived data remains an open question.

\textbf{Out-of-domain data.} It may be unreasonable to expect users to confine their queries to the intents a system supports, yet almost none of the datasets surveyed here include dedicated out-of-scope samples.
\emph{Clinc-150} and \emph{ROSTD} are the notable exceptions, though \emph{Clinc-150} is intent-only.
\citet{arora-etal-2024-intent} find that LLMs reject out-of-scope input more reliably when intent labels are fine-grained and the label space is small, so out-of-scope results from one dataset may not hold for another.
Practitioners can improvise---training on \emph{ATIS} and evaluating on \emph{Snips} yields pseudo-out-of-scope data---but we recommend that creators of new datasets include genuine out-of-scope queries.

\textbf{Multi-intent utterances.} Several surveyed datasets support multi-intent detection, where a single utterance carries more than one intent label: \emph{Task Oriented Parsing (TOP)}, \emph{MixATIS} and \emph{MixSNIPS}, \emph{NLU++}, \emph{BlendX}, \emph{CREDIT16}, and \emph{StanfordLU}.
The \emph{Mix} corpora join queries from a single parent using a small set of conjunctions, which limits their linguistic variety; \emph{BlendX} addresses this directly, and the difficulty it exposes suggests that the older benchmarks understate the task.
Joining utterances at random carries a further cost in realism: a user is unlikely to ask about the weather and rate a book in the same breath (\texttt{get\_weather} and \texttt{rate\_book}), so we encourage the creators of new datasets to collect multi-intent utterances rather than synthesize them.

\textbf{Challenge sets.} The near-ceiling scores reported by \citet{weld2021survey} might suggest that intent classification and slot-filling are close to solved; work on \emph{challenge} sets says otherwise.
\citet{larson-etal-2020-iterative} crowdsourced paraphrases of test samples from \emph{ATIS}, \emph{Snips}, \emph{Clinc-150}, and \emph{TOP}, and found that forbidding certain key words produced test data on which standard-trained models struggled.
Following \citet{ribeiro-etal-2020-beyond}, whose automated framework probes robustness in sentiment analysis and machine comprehension, \citet{peng2020etal-raddle}, \citet{liu2021robustness}, and \citet{sengupta-etal-2021-robustness} inject noise (spelling and ASR errors, speech disfluencies) into NLU data and apply modifications to it (paraphrases; punctuation, casing, and morphological changes).
Models trained on standard, clean data consistently degrade on these noisier inputs. We therefore encourage creators of new datasets to release challenge sets alongside their standard splits, so that robustness to paraphrase and noise is measured as a matter of course rather than retrofitted.

\subsection{Analysis of Datasets}

Our survey of intent classification and slot-filling corpora for task-oriented dialog systems has so far compared datasets using fairly straightforward attributes: the number of intents and slots, the number of utterances, language, and data source.
These catalog-level attributes, however, say little about the \emph{quality} of a benchmark.
Throughout this survey we have therefore supplemented them with several corpus-level diagnostics---computed uniformly across datasets (Section~\ref{sec:joint-datasets}), and including lexical variety measured by vocabulary size and type-token ratio \citep{token-type-ratio-ttr}---that begin to quantify three notions of quality: correctness, difficulty, and generalizability, which Table~\ref{tab:corpus-diagnostics} reports side by side.

\textbf{Correctness.} By correctness we mean the degree to which a corpus is free of annotation inconsistencies and errors \citep{larson-etal-2020-inconsistencies, klie-annotation-error-detection}.
Applying confident learning \citep{northcutt-2021-confident} and a semantic label-consistency check to the datasets we audit, we find that estimated label-noise rates vary widely: hand-curated English benchmarks such as \emph{Clinc-150}, \emph{Snips}, and \emph{xSID} sit below 1\%, whereas several crowd-translated multilingual partitions (e.g., of \emph{MASSIVE}) and datasets with heavily overlapping intents (e.g., \emph{HWU-64}) reach into the 10--17\% range.

Notably, almost none of the datasets surveyed here report inter-annotator agreement, so these post-hoc estimates are, for most corpora, the only available evidence of annotation reliability.
We caution that our estimates are encoder-dependent and thus comparative rather than absolute---for instance, the apparent noise in the Korean partition of \emph{MASSIVE} is partly an artifact of weaker multilingual representation---but the broad pattern, that derived and fine-grained corpora are noisier than carefully curated ones, is robust.

\textbf{Difficulty.} Recall the discussion of \emph{ATIS} and the notion of ``shallowness'' \citep{atis-shallow, niu-penn-2019-rationally-reappraising-atis}: high reported accuracy need not indicate a hard task.
As a uniform proxy for difficulty we report the accuracy of a fixed linear probe over sentence embeddings, alongside intent-distribution balance.
This exposes benchmarks that are near-saturated for reasons other than genuine ease---\emph{ATIS}, for example, pairs a highly imbalanced intent distribution (normalized entropy of 0.37) with essentially no embedding-space separation among its intents (a mean silhouette coefficient of $-0.05$)---as well as datasets whose fine-grained schemes make them genuinely hard (e.g., \emph{Leyzer} and \emph{Redwood}).

\textbf{Generalizability.} A test set can only measure generalization to the extent that it is not a near-copy of the training set.
Prior work has raised this concern---noting that models trained on these benchmarks often fail to generalize to real-world or ``long-tail'' inputs \citep{sengupta-etal-2021-robustness, larson-etal-2020-iterative, peng2020etal-raddle}, and proposing lower-overlap splits for the related task of semantic parsing \citep{finegan-dollak-etal-2018-improving}---and we quantify it directly by measuring, for each dataset, the fraction of test utterances that have a near-duplicate (embedding cosine above 0.90) in the training data.
The results are striking: over 79\% of \emph{Facebook}, 67\% of \emph{Banking-77}, and roughly 60\% of \emph{TOPv2} test utterances have such a near-duplicate in training, so a substantial share of their reported test performance reflects near-memorization rather than generalization.
Multi-intent corpora (e.g., \emph{MixSNIPS}) and out-of-scope-heavy test sets (e.g., \emph{HINT3}) exhibit the least overlap.

The purely lexical view agrees with these findings: the same benchmarks whose test sets are semantically redundant also present models with almost no unseen vocabulary, with token-level OOV rates below 1\% for \emph{Banking-77} (0.7\%), \emph{ATIS} (0.7\%), \emph{TOPv2} (0.8\%), and \emph{Facebook} (0.8\%).
At the other extreme sit the \emph{HINT3} corpora, whose test sets are drawn from deployed systems rather than from the same collection effort as their training data: between 28\% and 43\% of their test tokens are unseen in training, and roughly three quarters of their test word types are novel.
This gap is the clearest quantitative expression of a distinction that recurs throughout this survey. In some benchmarks the training and test splits are two samples of a single crowdsourcing effort; in others the test data independently reflect real usage.
Language typology matters as well: within the parallel partitions of \emph{MASSIVE}, which hold content constant, OOV rates track morphological richness, rising from 3.5\% for English through 5.7\% for German and 8.1\% for Russian to 11.3\% for Korean, so a fixed absolute accuracy is a harder result in a morphologically rich language than in English.

The audited set covers 37 of the 50 corpora surveyed here.\footnote{Of the remaining 13: \emph{MultiATIS++} is licence-restricted, inheriting the \textsc{ldc} terms of \emph{ATIS}; we could not locate an obtainable copy of \emph{CAIS}, \emph{FewJoint}, \emph{Almawave-SLU}, \emph{CSTOP} or \emph{StanfordLU}, which are variously distributed on request, released only in a few-shot episode format, or served from links that no longer resolve; \emph{NLU++}, \emph{ROSTD} and the smallest \emph{FoodOrdering} sets yield none of these measures; and the remainder are not covered by the present analysis. The gap reflects what we could obtain and process, not a judgement about these corpora.} Nor is every measure defined for every audited corpus: the intent-based measures require single-label intent annotations, and the test-set measures require a distributed test split.

\begin{table*}[tbp]
    \centering
    \footnotesize
    \begin{tabular}{llrrrrr}
        \toprule
        \textbf{Dataset} & \textbf{Task} & \textbf{Label noise} & \textbf{Probe acc.} & \textbf{Intent } $H$ & \textbf{Test OOV} & \textbf{Near-dup.} \\
        & & (\%) & (\%) & & (\%) & (\%) \\
        \midrule
        \emph{ask ubuntu} & Joint & 0.0 & 100 & 0.93 & 30.1 & 5 \\
        \emph{ATIS} & Joint & 2.1 & 93 & 0.37 & 0.7 & 43 \\
        \emph{BlendX (ATIS)} & Joint & --- & --- & --- & 2.6 & 9 \\
        \emph{BlendX (Banking-77)} & Joint & --- & --- & --- & 0.6 & 32 \\
        \emph{BlendX (Clinc-150)} & Joint & --- & --- & --- & 2.2 & 17 \\
        \emph{BlendX (Snips)} & Joint & --- & --- & --- & 5.5 & 4 \\
        \emph{Chatbot} & Joint & 1.0 & 94 & 0.99 & 9.9 & 32 \\
        \emph{Facebook (en)} & Joint & 0.2 & 98 & 0.69 & 0.8 & 79 \\
        \emph{Facebook (es)} & Joint & 0.8 & 94 & 0.74 & 3.8 & 64 \\
        \emph{Facebook (th)} & Joint & 1.0 & 94 & 0.79 & 0.1 & 66 \\
        \emph{HWU-64} & Joint & 13.8 & 85 & 0.92 & --- & --- \\
        \emph{Leyzer (en)} & Joint & 8.4 & 77 & 0.85 & 8.0 & 23 \\
        \emph{Leyzer (es)} & Joint & 9.2 & 79 & 0.85 & 10.2 & 61 \\
        \emph{Leyzer (pl)} & Joint & 9.2 & 77 & 0.83 & 11.7 & 55 \\
        \emph{MASSIVE (ar)} & Joint & 17.1 & 65 & 0.92 & 9.2 & 37 \\
        \emph{MASSIVE (de)} & Joint & 13.2 & 71 & 0.92 & 5.7 & 41 \\
        \emph{MASSIVE (en)} & Joint & 5.9 & 83 & 0.92 & 3.5 & 32 \\
        \emph{MASSIVE (es)} & Joint & 8.6 & 77 & 0.92 & 4.4 & 39 \\
        \emph{MASSIVE (he)} & Joint & 11.2 & 73 & 0.92 & 7.4 & 37 \\
        \emph{MASSIVE (hi)} & Joint & 8.8 & 77 & 0.92 & 3.8 & 35 \\
        \emph{MASSIVE (ja)} & Joint & 9.0 & 78 & 0.92 & 0.4 & 44 \\
        \emph{MASSIVE (ko)} & Joint & 16.6 & 63 & 0.92 & 11.3 & 78 \\
        \emph{MASSIVE (ru)} & Joint & 8.6 & 77 & 0.92 & 8.1 & 43 \\
        \emph{MASSIVE (zh)} & Joint & 8.0 & 79 & 0.92 & 0.5 & 45 \\
        \emph{MixATIS} & Joint & --- & --- & --- & 3.1 & 7 \\
        \emph{MixSNIPS} & Joint & --- & --- & --- & 5.4 & 3 \\
        \emph{MTOP (de)} & Joint & 9.7 & 81 & 0.74 & 5.1 & 38 \\
        \emph{MTOP (en)} & Joint & 4.5 & 86 & 0.74 & 3.5 & 31 \\
        \emph{MTOP (es)} & Joint & 7.8 & 83 & 0.74 & 4.6 & 35 \\
        \emph{MTOP (fr)} & Joint & 9.2 & 80 & 0.75 & 4.4 & 33 \\
        \emph{MTOP (hi)} & Joint & 9.1 & 80 & 0.75 & 4.8 & 37 \\
        \emph{MTOP (th)} & Joint & 9.2 & 81 & 0.75 & 0.3 & 37 \\
        \emph{PhoATIS} & Joint & --- & --- & --- & 1.6 & 53 \\
        \emph{Snips} & Joint & 0.4 & 97 & 1.00 & 5.4 & 17 \\
        \emph{TOP} & Joint & 4.9 & 90 & 0.67 & 1.5 & 38 \\
        \emph{TOPv2} & Joint & 6.8 & 88 & 0.76 & 0.8 & 60 \\
        \emph{Web Applications} & Joint & 0.0 & 95 & 0.91 & 40.8 & 2 \\
        \emph{xSID} & Joint & 0.2 & 97 & 0.78 & 3.4 & 53 \\
        \midrule
        \emph{ACID} & Intent & 3.7 & 87 & 0.92 & 1.4 & 59 \\
        \emph{Banking-77} & Intent & 2.3 & 88 & 0.99 & 0.7 & 67 \\
        \emph{Clinc-150} & Intent & 0.6 & 94 & 1.00 & 4.8 & 36 \\
        \emph{CREDIT16} & Intent & --- & --- & --- & 2.6 & 0 \\
        \emph{Curekart} & Intent & 6.3 & 76 & 0.89 & 28.3 & 8 \\
        \emph{MInDS-14} & Intent & 0.0 & 97 & 1.00 & --- & --- \\
        \emph{Outlier (Random)} & Intent & 0.1 & 99 & 1.00 & --- & --- \\
        \emph{Outlier (Same)} & Intent & 0.1 & 99 & 1.00 & --- & --- \\
        \emph{Outlier (Unique)} & Intent & 0.1 & 98 & 1.00 & --- & --- \\
        \emph{Powerplay11} & Intent & 3.2 & 76 & 0.92 & 43.0 & 4 \\
        \emph{Redwood} & Intent & 5.6 & 81 & 1.00 & 1.2 & 56 \\
        \emph{SOFMattress} & Intent & 3.0 & 84 & 0.97 & 32.2 & 7 \\
        \midrule
        \emph{Airbnb} & Slot & --- & --- & --- & 2.9 & 20 \\
        \emph{Burrito} & Slot & --- & --- & --- & 6.0 & 18 \\
        \emph{Greyhound} & Slot & --- & --- & --- & 2.8 & 43 \\
        \emph{MIT Movie} & Slot & --- & --- & --- & 3.3 & 9 \\
        \emph{MIT Movie (trivia10k13)} & Slot & --- & --- & --- & 3.2 & 5 \\
        \emph{MIT Restaurant} & Slot & --- & --- & --- & 2.8 & 26 \\
        \emph{OpenTable} & Slot & --- & --- & --- & 4.5 & 18 \\
        \emph{Pizza} & Slot & --- & --- & --- & 1.8 & 31 \\
        \emph{Restaurants-8k} & Slot & --- & --- & --- & 4.2 & 45 \\
        \emph{Sub} & Slot & --- & --- & --- & 9.1 & 5 \\
        \emph{United} & Slot & --- & --- & --- & 0.4 & 44 \\
        \bottomrule
    \end{tabular}
    \caption{Corpus-level diagnostics for the 37 corpora we could load and audit; language partitions and collection variants are listed separately, giving 61 rows. Grouped by task, alphabetical within each group. \emph{Label noise}: share of training utterances flagged by confident learning. \emph{Probe acc.}: five-fold linear-probe accuracy over sentence embeddings. \emph{Intent }$H$: normalized intent entropy, 1.00 being uniform. \emph{Test OOV}: share of test tokens unseen in training. \emph{Near-dup.}: share of test utterances whose nearest training utterance exceeds cosine 0.90. \emph{Task}, and the intent and slot counts, follow the catalog tables of Sections~\ref{sec:joint-datasets} to~\ref{sec:slot-filling-datasets}. A dash marks a measure that does not apply: the intent-based measures need single-label intent annotations, the test-set measures a distributed test split.}
    \label{tab:corpus-diagnostics}
\end{table*}

\subsection{Slot Annotation Schemes}\label{subsec:slot-schemes}
The slot-filling half of the intent-classification/slot-filling duo receives markedly less analytical attention than the intent half---in this survey and in the literature more broadly---yet slot schemes differ in ways that materially affect cross-dataset comparability.
We highlight one such axis here.

Two broad philosophies govern how datasets name their slots.
\emph{Role-based} schemes label a value by the role it plays: \emph{ATIS}, for instance, distinguishes \texttt{fromloc.city\_name}, \texttt{toloc.city\_name}, and \texttt{stoploc.city\_name}, so that the same city is tagged differently depending on whether it is an origin, destination, or stopover.
\emph{Entity-typed} schemes instead label a value by what it \emph{is} rather than by the role it fills, and so do not separate, say, an origin city from a destination. \emph{Snips} is the clearest case in this survey: almost no value in it appears under more than one slot type.
Role-based schemes capture the structure a downstream system needs, but they proliferate slot types; entity-typed schemes are simpler, but they conflate roles. \emph{SLURP} \citep{bastianelli-etal-2020-slurp}, for example, labels both currencies in \emph{``what is one American dollar in Japanese yen''} with the same slot type, leaving no way to distinguish source from target.

We quantify where a dataset sits on this axis with a measure we call \emph{value polysemy}: the fraction of distinct slot values that appear under more than one slot type.
Role-based schemes have high polysemy by construction (a city recurs across origin, destination, and stopover roles), whereas entity-typed schemes have low polysemy (a value maps to a single type).
Table~\ref{tab:slot-schemes} reports this for the 19 corpora whose slot annotations we were able to extract.
The contrast is stark: 30\% of \emph{ATIS} slot values are polysemous, versus 1.1\% for \emph{Snips}.
\emph{TOP} shows that the measure catches role structure even when the labels do not advertise it. It uses no \emph{ATIS}-style dotted role names, yet its 13\% polysemy follows from slots such as \texttt{source}, \texttt{destination}, and \texttt{waypoint} sharing the same city values.
Datasets derived from a parent inherit its scheme: \emph{MixATIS} (27\%) tracks \emph{ATIS}, and \emph{MixSNIPS} (1.1\%) matches \emph{Snips} exactly.

Not all polysemy is deliberate role structure, however.
\emph{Snips}'s small residual comes from genuinely ambiguous bare numerics (``four'' is annotated as \texttt{party\_size\_number}, \texttt{rating\_value}, and \texttt{timeRange} in different utterances), and in \emph{MIT-Restaurant} a value like ``bar'' is spread across five slot types (\texttt{Amenity}, \texttt{Cuisine}, \texttt{Location}, \texttt{Price}, \texttt{Restaurant\_Name}) in a way that reads more like annotation inconsistency than a designed distinction.
Value polysemy thus serves double duty: as a descriptor of schema style, and---for entity-typed datasets, where it should be near zero---as a pointer to slot-label inconsistency.

A second axis of slot-scheme variation, orthogonal to role-marking, is the diversity of the \emph{values} each slot takes.
Distinguishing \emph{open-class} slots (whose fillers are effectively unbounded, such as song or place names) from \emph{closed-class} slots (whose fillers are enumerable, such as party size or a yes/no flag) matters because open-class slots are where models face out-of-vocabulary fillers and struggle to generalize---the slot-level counterpart of the lexical-generalization concern we raised for intents.
This axis is independent of schema style, and the two together clarify what makes a slot dataset hard.
\emph{ATIS} is the telling case: although it has the most role-marked slot types, most are closed-class (36 of its slots take five or fewer distinct values), and even its ``open'' slots draw on a small world---only 57 distinct cities fill \texttt{toloc.city\_name} across more than 4{,}000 occurrences---giving a slot-filler type-token ratio of just 0.04.
\emph{Snips} and the \emph{MIT} corpora sit at the opposite corner: they have fewer and structurally simpler (entity-typed, low-polysemy) slots, but those slots are overwhelmingly open-class and high-cardinality---\emph{Snips}'s \texttt{object\_name} takes 2{,}822 distinct values in 2{,}936 occurrences, and nearly every \emph{MIT-Restaurant} slot is open---for type-token ratios around 0.3.
In short, \emph{ATIS} is hard in its \emph{role} structure but lexically shallow, whereas \emph{Snips} and \emph{MIT} are structurally simple but lexically open. Reporting one slot-filling F1 for both says nothing about which of the two difficulties a model has actually handled.

A third property bears directly on overfitting risk: how predictable a slot's surrounding \emph{context} is.
\citet{larson-etal-2020-data} observed that \emph{ATIS} slots are heavily cued by adjacent tokens---origin and destination cities almost always follow the prepositions ``from'' and ``to''.
We make this measurable for every slot type as the share of occurrences in which its single most common neighboring token recurs (the \emph{Ctx.\ pred.}\ column of Table~\ref{tab:slot-schemes} averages this over a dataset's slots).
The concern is that a high share lets a model locate and label a slot from a fixed trigger word rather than from the slot's content, inflating apparent performance.
The effect is large and uneven.
In \emph{ATIS}, \texttt{toloc.city\_name} is preceded by ``to'' in 85\% of its occurrences and \texttt{fromloc.city\_name} by ``from'' in 83\% (and is itself followed by ``to'' in 80\%), so an \emph{ATIS} slot's modal left neighbor covers 66\% of occurrences on average, and 35 of its slot types have a neighbor that predicts them at least 70\% of the time.
\emph{MixATIS} inherits this cueing (61\%), and the grammar-generated \emph{Leyzer} is similarly templated (50\%).
By contrast, the \emph{MIT} corpora have no dominant triggers (a modal left-neighbor share of just 22\%, and not a single trigger-predictable slot), which their \emph{free-form} collection strategy would lead one to expect: workers wrote queries without being given slot values to build them around.
A slot-filling result on \emph{ATIS} therefore reflects, in part, a model's ability to exploit a handful of function words, whereas the same score on \emph{MIT-Restaurant} demands recognizing slots in genuinely diverse contexts.
The nested \emph{MTOP} is likewise strongly cued (55\%), while its predecessor \emph{TOP} and the multilingual \emph{xSID} and \emph{MASSIVE} are markedly lower.
Across \emph{MASSIVE}'s languages the slot scheme is essentially constant---55 entity-typed slot types with roughly 5\% value polysemy in every locale---but context predictability varies with word order, from 22\% in German to 55\% in Japanese, in part reflecting the tokenization of scripts that are not space-delimited.

The broader point is that a slot-filling F1 score is not directly comparable across these schemes.
Reaching high F1 on \emph{ATIS}'s 79 role-marked types is a different task than on \emph{MIT-Restaurant}'s 8 entity types, and reported numbers should be read with the annotation scheme in mind.
Systematically cataloging slot schemes is a promising direction for making slot-filling evaluation as rigorous as intent-classification evaluation, as is auditing entity-typed datasets for the inconsistency that low-but-nonzero polysemy points to.

\begin{table*}[t]
\centering
\scalebox{0.785}{
\begin{tabular}{lrrrrrl}
\toprule
\textbf{Dataset} & \textbf{Slot types} & \textbf{Slots/utt} & \textbf{Value polysemy} & \textbf{Open\,/\,closed} & \textbf{Ctx.\ pred.} & \textbf{Scheme} \\
\midrule
\emph{ATIS} & 79 & 3.3 & 30.0\% & 5\,/\,36 & 66\% & role-based \\
\emph{MixATIS} & 74 & 5.2 & 27.3\% & 3\,/\,36 & 61\% & role-based \\
\emph{United} & 13 & 3.2 & 22.2\% & 9\,/\,1 & 65\% & role-based (implicit) \\
\emph{Greyhound} & 13 & 1.0 & 13.8\% & 4\,/\,1 & 36\% & role-based (implicit) \\
\emph{TOP} & 35 & 2.0 & 13.2\% & 21\,/\,4 & 38\% & role-based (implicit) \\
\emph{TOPv2} & 81 & 1.8 & 9.1\% & 43\,/\,14 & 32\% & entity-typed \\
\emph{Airbnb} & 11 & 0.9 & 7.4\% & 4\,/\,0 & 31\% & entity-typed \\
\emph{Leyzer (en)} & 87 & 4.4 & 6.8\% & 26\,/\,33 & 50\% & entity-typed \\
\emph{MTOP} & 77 & 1.7 & 6.2\% & 30\,/\,15 & 55\% & entity-typed \\
\emph{MASSIVE (hi)} & 55 & 1.0 & 5.1\% & 23\,/\,4 & 34\% & entity-typed \\
\emph{MASSIVE (de)} & 55 & 1.0 & 5.0\% & 24\,/\,5 & 22\% & entity-typed \\
\emph{MASSIVE (ja)} & 55 & 1.0 & 5.0\% & 23\,/\,6 & 55\% & entity-typed \\
\emph{MASSIVE (en)} & 55 & 1.0 & 4.6\% & 22\,/\,7 & 26\% & entity-typed \\
\emph{MIT-Restaurant} & 8 & 2.0 & 4.2\% & 8\,/\,0 & 22\% & entity-typed \\
\emph{MIT-Movie} & 12 & 2.2 & 3.6\% & 10\,/\,0 & 30\% & entity-typed \\
\emph{xSID} & 41 & 1.9 & 1.4\% & 18\,/\,9 & 34\% & entity-typed \\
\emph{Snips} & 39 & 2.6 & 1.1\% & 20\,/\,4 & 43\% & entity-typed \\
\emph{MixSNIPS} & 39 & 5.2 & 1.1\% & 20\,/\,4 & 43\% & entity-typed \\
\emph{OpenTable} & 6 & 1.0 & 0.0\% & 4\,/\,0 & 26\% & entity-typed \\
\bottomrule
\end{tabular}}
\caption{Slot annotation schemes across the 19 corpora whose slot spans we extract. \emph{Value polysemy} is the fraction of distinct slot values appearing under more than one slot type, and is high for role-based schemes and low for entity-typed ones. \emph{Open\,/\,closed} counts slot types taking $\geq$50 versus $\leq$5 distinct filler values. \emph{Ctx.\ pred.}\ is the occurrence-weighted mean share of each slot's most common left-neighbor token, a proxy for reliance on trigger words rather than slot content. Schemes marked \emph{implicit} encode role in the slot name without \emph{ATIS}-style dotted notation. Derived datasets (\emph{MixATIS}, \emph{MixSNIPS}) inherit their parents' schemes.}
\label{tab:slot-schemes}
\end{table*}

\subsection{Intent Classification and Slot-Filling in the Era of Large Language Models}\label{subsec:llm-era}
The landscape of natural language understanding has shifted substantially in recent years.
Large language models (LLMs) such as GPT-4 and the Llama family can perform intent classification and slot-filling in a zero-shot or few-shot setting, without any task-specific training data.
This raises a natural question: do the benchmark datasets cataloged in this survey remain relevant?

We argue that they do, and that the analyses in this survey are what make them useful in this setting, by indicating \emph{which} datasets remain diagnostic and \emph{why}.
First, these corpora provide the held-out, human-labeled test sets against which LLM NLU capabilities can be rigorously \emph{measured}; without principled benchmarks---documented annotation guidelines, standardized protocols, and gold labels---claims about LLM performance rest on anecdote.
Recent few-shot and in-context LLM methods for intent detection and discovery are, accordingly, benchmarked directly on the corpora surveyed here, such as \emph{Clinc-150} and \emph{Banking-77} \citep{sahu-etal-2022-data, rodriguez-2024-intentgpt}.
Not every dataset is equally informative in the LLM era, and the corpus-level diagnostics developed above help narrow the field: our measures of train/test redundancy, saturation, and label noise indicate which benchmarks still discriminate between systems (the harder, less redundant, cleaner ones) and which now return near-ceiling scores that reveal little.
\citet{rodrigues-vas-2026-when} show what is at stake in that choice: pitting a fine-tuned RoBERTa against a zero-shot LLM, they find the verdict depends on which benchmark is used---on \emph{ATIS} the supervised model wins by 11.8 accuracy points (95.9 versus 84.1), whereas on the 150-intent \emph{Clinc-150} schema the two are statistically indistinguishable (89.1 versus 88.5)---so the same experiment run on either corpus alone would support the opposite conclusion about whether LLMs have displaced task-specific classifiers.

Second, the two tasks fare very differently under LLMs, a gap that only the slot-filling corpora make visible.
Intent classification is a text-classification problem that LLMs handle well zero-shot, but faithful slot-filling---extracting spans that conform to a fixed schema---remains difficult: \citet{rana-etal-2025-zero} report that off-the-shelf LLMs are markedly weaker at zero-shot slot-filling and require substantial adaptation to become competitive.

Third, the multilingual and low-resource datasets cataloged here are especially important for evaluating LLM capabilities beyond English, where transfer degrades for lower-resource languages and under-represented scripts.
Datasets such as \emph{MASSIVE}, \emph{MultiATIS++}, and \emph{MInDS-14} provide the parallel infrastructure needed to quantify these gaps.

Fourth, LLMs have increasingly become \emph{data generation tools} for intent and slot datasets---as in \emph{BlendX} \citep{yoon-etal-2024-blendx}, which uses ChatGPT to synthesize blended multi-intent utterances, and in prompting-based augmentation for intent classification \citep{sahu-etal-2022-data}.
Tellingly, \citet{sahu-etal-2022-data} find that such generation breaks down on fine-grained label sets, where the model produces utterances belonging to a closely related intent---the same fine-grained confusability we quantify for \emph{Banking-77} and \emph{HWU-64}---so the human-labeled originals remain the ground truth these generators are checked against.
This reliance on LLMs also introduces new concerns about benchmark contamination (the risk that LLMs have trained on benchmark test sets) and label reliability that must be taken seriously when interpreting LLM evaluation results.
A further open problem is evaluation methodology itself: unlike a classifier's argmax over a fixed label set, a generative model's free-form output must be mapped back to gold intents and slot spans, and how one scores that mapping (exact match, schema conformance, or semantic equivalence) can materially change reported performance---an area the community has yet to standardize.

Benchmark contamination deserves particular emphasis, because it interacts with the popularity of the very datasets this survey recommends.
Long-standing benchmarks such as \emph{Snips}, \emph{Clinc-150}, and \emph{Banking-77} have been freely available on the public web for years, often on GitHub and the Hugging Face Hub with their gold labels, and \emph{ATIS}, though formally \textsc{ldc}-licensed, has been redistributed just as widely. All are therefore likely to appear, in whole or in part, in the web-scraped corpora used to pretrain modern LLMs.
When this is the case, a strong zero-shot score cannot be cleanly attributed to genuine language understanding rather than memorization of the test set.
Contamination compounds the train/test overlap documented above: for a dataset in which the majority of test utterances already have near-duplicates in their own training split (for example, \emph{Banking-77} or \emph{Facebook}), an LLM that has encountered the corpus during pretraining is doubly unlikely to be generalizing.
We therefore recommend that LLM evaluations report, wherever possible, whether the benchmark predates the model's training cutoff, prefer newer or privately held-out test sets when contamination is plausible, and treat near-ceiling zero-shot results on classic benchmarks with corresponding caution.
We note that contamination is difficult to verify for closed models whose training data is undisclosed, which is itself an argument for maintaining fresh, well-documented evaluation corpora.

In short, the benchmarks in this survey remain the backbone of rigorous evaluation for structured NLU.
As powerful but opaque LLMs enter the dialog system development pipeline, the need for transparent, reproducible, and diverse evaluation corpora only grows.

\section{Conclusion}\label{sec:conclusion}
The past decade has brought remarkable growth in the breadth and diversity of intent classification and slot-filling datasets, and in the sophistication of the models evaluated on them.
The emergence of large language models (LLMs) as capable zero-shot NLU systems has further transformed the landscape, but it has not made task-specific benchmarks obsolete.
Without principled, human-labeled benchmarks covering diverse domains, languages, and annotation schemes, the field cannot make credible distinctions between genuine progress in NLU and dataset-specific overfitting or prompt engineering.

In this survey, we have cataloged corpora for intent classification, slot-filling, and joint modeling spanning a wide range of languages, domains, and collection methodologies.
Our survey reveals several persistent gaps that remain important targets for future dataset development: the dominance of English and a small set of high-resource languages; the relative scarcity of realistic multi-intent data; the underrepresentation of spoken and noisy inputs; and the limited availability of data from real user interactions as opposed to crowdsourced elicitations.
We echo prior recommendations that new datasets address these gaps, and we add that the LLM era introduces new challenges around benchmark contamination---the risk that LLMs have been trained on benchmark test sets---that must be taken seriously when designing evaluations.

Our hope is that this survey serves as a practical reference for researchers and practitioners selecting, comparing, or building upon existing datasets, and as a call to action for the development of richer, more diverse, and more realistic benchmarks for the task-oriented dialog systems of the future.

%
%
%

\impact{This survey catalogs publicly available datasets for intent classification and slot-filling in task-oriented dialog systems. We do not anticipate direct negative societal impacts from this work. By consolidating and making accessible a broad set of evaluation benchmarks, we hope to support more rigorous and reproducible NLU research, including evaluation of large language models on structured tasks. Researchers should be aware that many datasets were collected via crowdsourcing and may reflect demographic and linguistic biases of the annotator pools used.}

\acks{We thank Deborah Dahl for granting us an interview about the origins and development of \emph{ATIS}; her recollections informed our account of that corpus.}


\section*{Declaration of generative AI and AI-assisted technologies in the writing process}
During the preparation of this work the authors used Anthropic's Claude in order to draft and revise portions of the manuscript text, to summarize and describe the results of the quantitative dataset analyses reported in Section~\ref{sec:discussion-future-work}, and to assist with reference checking and bibliography maintenance. After using this tool, the authors reviewed and edited the content as needed and take full responsibility for the content of the published article.

\bibliographystyle{cas-model2-names}
\bibliography{compling_style_cas}

\end{document}